\documentclass[11pt]{article}

\usepackage[final]{acl}
\usepackage{times}
\usepackage{latexsym}
\usepackage[T1]{fontenc}
\usepackage[utf8]{inputenc}
\usepackage{microtype}
\usepackage{inconsolata}
\usepackage{graphicx}
\usepackage{tcolorbox}
\usepackage{listings}
\usepackage{xcolor}
\usepackage{tikz}
\usetikzlibrary{arrows.meta, positioning, decorations.pathreplacing, calc, fit}
\usepackage{amsmath}
\usepackage{amssymb}
\usepackage{enumitem}
\usepackage{caption}
\definecolor{palepurple}{RGB}{220,200,225}
\usepackage{booktabs,multirow,array}
\usepackage{algorithm}
\usepackage{algpseudocode}

\lstdefinelanguage{json}{
    basicstyle=\ttfamily\small,
    numbers=none,
    showstringspaces=false,
    breaklines=true,
    frame=none
}

\title{AILS-NTUA at SemEval-2026 Task 12: Graph-Based Retrieval and Reflective Prompting for Abductive Event Reasoning}

\author{
    Nikolas Karafyllis, Maria Lymperaiou, Giorgos Filandrianos,  \\  \textbf{Athanasios Voulodimos, Giorgos Stamou} \\ 
    School of Electrical and Computer Engineering,  AILS Laboratory\\
    National Technical University of Athens \\
    \texttt{\href{mailto:nickarafyllis@gmail.com}{nickarafyllis@gmail.com}, \{\href{mailto:marialymp@ails.ece.ntua.gr}{marialymp}, \href{mailto:geofila@ails.ece.ntua.gr}{geofila}\}@ails.ece.ntua.gr, } \\
    \texttt{\href{mailto:thanosv@mail.ntua.gr}{thanosv@mail.ntua.gr},
    \href{mailto:gstam@cs.ntua.gr}{gstam@cs.ntua.gr}}\\
}

\begin{document}
\maketitle
\begin{abstract}
We present a winning three-stage system for SemEval 2026 Task~12: Abductive Event Reasoning that combines graph-based retrieval, LLM-driven abductive reasoning with prompt design optimized through reflective prompt evolution, and post-hoc consistency enforcement; our system ranks first on the evaluation-phase leaderboard with an accuracy score of 0.95.
Cross-model error analysis across 14 models (7~families) reveals three shared inductive biases: causal chain incompleteness, proximate cause preference, and salience bias, whose cross-family convergence (51\% cause-count reduction) indicates systematic rather than model-specific failure modes in multi-label causal reasoning.

\end{abstract}

\section{Introduction}
In a rapidly evolving digital world, information travels at an unprecedented scale, resulting in a constant flood of events. Such dissemination of information does not occur in isolation; instead, it triggers interconnected event chains in which each specific occurrence can act as the root cause of subsequent events. Thus, understanding \textit{why} an event happens within context becomes critical, requiring the discrimination of true cause-and-effect relationships from mere correlations.

Large Language Models (LLMs) have demonstrated great potential in event extraction and prediction \cite{chang2025comprehensiveevaluationlargelanguage, tanev-etal-2025-exploring, srivastava-etal-2025-instruction, li2025eventextractionlargelanguage}. Still, this ability does not account for \textit{abductive reasoning}, which concerns inferring the most plausible cause from incomplete information. Naturally, abduction explains why an event most possibly happened within a causal chain, offering valuable insights and interpretations for chains of events.

A large body of research focuses on LLM reasoning \cite{giadikiaroglou-etal-2024-puzzle, ke2025a, liu2025logicalreasoninglargelanguage}, yet abductive reasoning remains particularly challenging, since it engages parallel streams of thinking and implicit causal attribution rather than applying deterministic inference rules. For this reason, it serves as a stress test for LLMs, exposing limitations that are less apparent in hypothesis-free reasoning settings \cite{logical-reasoners, floridi2025kindreasoningifany, he2025from}. 

Motivated by the 
challenges entailed within real-world, uncertainty-rich data setups, SemEval 2026 Task 12 \textbf{Abductive Event Reasoning: Towards Real-World Event Causal Inference for Large Language Models} ~\cite{cao2026semeval} investigates LLMs' abilities in event-level causal reasoning: given an event and a set of retrieved documents, the LLM is tasked with identifying the most direct and plausible cause and forming the best explanation by harnessing prior knowledge, context understanding and hypothesis justification. 
In our work, we address this task through a three-stage pipeline 
that \textbf{ranks first on the evaluation leaderboard} scoring at 0.95/1.00.
Overall, our contributions are:
\begin{itemize}[noitemsep,topsep=2pt]
    \item We evaluate 18 model configurations across seven families, characterizing the performance landscape and persistent challenges of frontier LLMs on abductive event reasoning.
    \item We perform a thorough error analysis across 14 models (7~families), identifying three shared inductive biases: causal chain incompleteness, proximate cause preference, and salience bias, that produce conservative cause selection as the dominant failure mode.
\end{itemize}

The code for our system is available on GitHub\footnote{https://github.com/nickarafyllis/semeval-2026-task12}.

\section{Background}
\paragraph{Task description} The task targets the selection of the most direct, plausible cause of a real-world event, as evidenced via a \textit{single-step} reasoning process based on the given textual evidence. Each instance comprises an \textit{event}, \textit{context} and four \textit{candidate explanation} options, framing abductive reasoning within a multiple-choice question-answering setup. The \textit{event} field contains a textual description of a real-world event; \textit{context} refers to a set of documents that may either be related to the event or act as distractors; finally, one or more \textit{candidate explanations} may be correct given the input context, while an option stating that ``None of the causes is correct'' is always included. An LLM is evaluated on its ability to reason over the context for an event by selecting suitable candidate explanations. A full match between predicted and ground truth responses yields 1 point, a partial match earns 0.5 points, and a wrong/invalid selection receives 0 points.

\paragraph{Related work} Abductive reasoning with LLMs has received some notable contributions in recent years. First benchmarks, such as aNLI, established the field by requiring the selection of the best explanatory bridge between observations~\cite{bhagavatula2020abductive}. With the rise of LLMs, generation of plausible hypotheses is feasible but often inconsistent, especially under ambiguous evidence~\cite{pareschi2023abductivereasoninggpt4language}. Genuine abduction may also be obscured by dataset artifacts, which can be compromised via appropriate prompting~\cite{balepur-etal-2024-artifacts}. In parallel, abduction is reframed as verifiable reasoning, promoting faithful hypothesis evaluation and robustness to noise~\cite{he2024causejudgeridentifyingcausellms}. Additional work evaluates abduction in more formal logical settings (e.g., syllogistic forms), revealing systematic biases and limitations that persist even in advanced LLMs~\cite{abe-syllogistic}. At the same time, works emphasizing abduction as a stress test for LLMs shows that generating explanations under uncertainty is particularly challenging compared to other forms of reasoning~\cite{logical-reasoners}, while evaluating explanations within diverse viewpoints provides insightful probing for LLM failures~\cite{he2025geargeneralevaluationframework}. In total, robust abductive inference persists as a key challenge, despite promising evidence arising from prompting and task design~\cite{dougrez-lewis-etal-2025-assessing}.


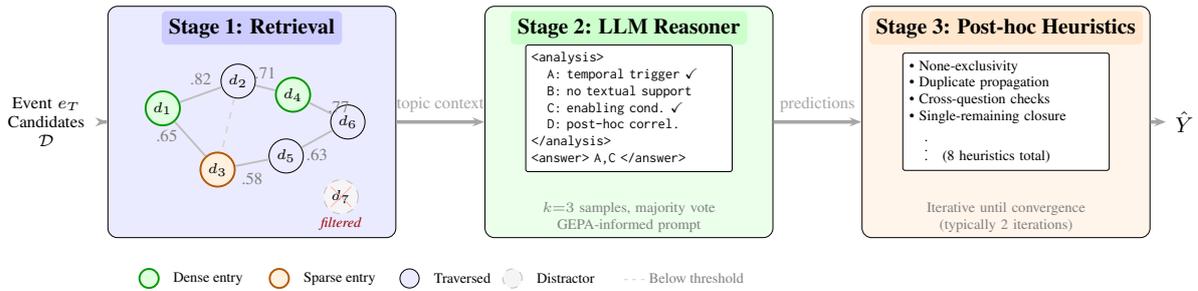
\begin{figure*}[t!]
\vskip -0.15in
\centering
\resizebox{\textwidth}{!}{%
\begin{tikzpicture}[
    stage/.style={draw, rounded corners=4pt, fill=#1, minimum width=4.2cm,
                  minimum height=3.4cm, inner sep=6pt},
    stagelabel/.style={font=\small\bfseries, anchor=north},
    doc/.style={draw, circle, fill=blue!8, minimum size=0.5cm,
                font=\tiny, inner sep=0pt},
    entrydense/.style={draw=green!60!black, circle, fill=green!12, minimum size=0.5cm,
                       font=\tiny, inner sep=0pt, line width=0.8pt},
    entrysparse/.style={draw=orange!70!black, circle, fill=orange!12, minimum size=0.5cm,
                        font=\tiny, inner sep=0pt, line width=0.8pt},
    distractor/.style={draw=gray!50, circle, fill=gray!8, minimum size=0.5cm,
                       font=\tiny, inner sep=0pt, dashed},
    xmlblock/.style={draw, rounded corners=1pt, fill=white,
                     font=\tiny\ttfamily, inner sep=3pt, text width=2.8cm},
    rulebox/.style={draw, rounded corners=1pt, fill=white,
                    font=\tiny, inner sep=3pt, text width=2.8cm},
    bigarrow/.style={-{Stealth[length=5pt]}, line width=1.5pt, gray!70},
    edge/.style={thick, gray!50},
    annot/.style={font=\tiny, gray},
    >=Stealth
]

\node[stage=blue!8] (s1) at (0, 0) {};
\node[stagelabel, fill=blue!20, rounded corners=2pt, inner sep=3pt]
    at ([yshift=-2pt]s1.north) {Stage 1: Retrieval};

\node[entrydense] (d1) at (-1.3, 0.2) {$d_1$};
\node[entrydense] (d4) at (0.6, 0.4) {$d_4$};
\node[entrysparse] (d3) at (-0.5, -0.7) {$d_3$};
\node[doc] (d2) at (-0.2, 0.6) {$d_2$};
\node[doc] (d5) at (0.5, -0.5) {$d_5$};
\node[doc] (d6) at (1.4, 0.0) {$d_6$};
\node[distractor] (d7) at (1.3, -1.1) {$d_7$};

\draw[edge] (d1) -- node[above, font=\scriptsize, gray] {.82} (d2);
\draw[edge] (d1) -- node[left, font=\scriptsize, gray, pos=0.4] {.65} (d3);
\draw[edge] (d2) -- node[above, font=\scriptsize, gray] {.71} (d4);
\draw[edge] (d3) -- node[below, font=\scriptsize, gray] {.58} (d5);
\draw[edge] (d4) -- node[right, font=\scriptsize, gray, pos=0.4] {.77} (d6);
\draw[edge] (d5) -- node[below, font=\scriptsize, gray] {.63} (d6);
\draw[edge, dashed, gray!30] (d2) -- (d3);

\draw[red!30] ($(d7)+(-0.15,-0.15)$) -- ($(d7)+(0.15,0.15)$);
\draw[red!30] ($(d7)+(-0.15,0.15)$) -- ($(d7)+(0.15,-0.15)$);
\node[annot, red!60!black, font=\tiny\itshape] at (1.3, -1.5) {filtered};


\node[stage=green!8] (s2) at (5.5, 0) {};
\node[stagelabel, fill=green!20, rounded corners=2pt, inner sep=3pt]
    at ([yshift=-2pt]s2.north) {Stage 2: LLM Reasoner};

\node[xmlblock] at (5.5, 0.2) {%
    \textless analysis\textgreater\\
    \quad A: temporal trigger \checkmark\\
    \quad B: no textual support\\
    \quad C: enabling cond.\ \checkmark\\
    \quad D: post-hoc correl.\\
    \textless /analysis\textgreater\\
    \textless answer\textgreater\ A,C \textless /answer\textgreater};

\node[annot, text width=3.2cm, align=center] at (5.5, -1.4)
    {$k{=}3$ samples, majority vote\\GEPA-informed prompt};

\node[stage=orange!8] (s3) at (11.0, 0) {};
\node[stagelabel, fill=orange!20, rounded corners=2pt, inner sep=3pt]
    at ([yshift=-2pt]s3.north) {Stage 3: Post-hoc Heuristics};

\node[rulebox] at (11.0, 0.15) {%
    \textbullet\ None-exclusivity\\
    \textbullet\ Duplicate propagation\\
    \textbullet\ Cross-question checks\\
    \textbullet\ Single-remaining closure\\
    \quad\vdots\quad (8 heuristics total)};

\node[annot, text width=2.8cm, align=center] at (11.0, -1.4)
    {Iterative until convergence\\(typically 2 iterations)};

\draw[bigarrow] (s1.east) -- node[above, font=\scriptsize] {topic context} (s2.west);
\draw[bigarrow] (s2.east) -- node[above, font=\scriptsize] {predictions} (s3.west);

\node[font=\scriptsize, text width=1.5cm, align=center] (input) at (-3.0, 0)
    {Event $e_T$\\Candidates\\$\mathcal{D}$};
\draw[bigarrow] (input.east) -- (s1.west);

\node[font=\small\bfseries] (output) at (13.6, 0) {$\hat{Y}$};
\draw[bigarrow] (s3.east) -- (output.west);

\node[entrydense, minimum size=0.28cm] at (-1.5, -2.3) {};
\node[font=\tiny, right] at (-1.28, -2.3) {Dense entry};
\node[entrysparse, minimum size=0.28cm] at (0.4, -2.3) {};
\node[font=\tiny, right] at (0.62, -2.3) {Sparse entry};
\node[doc, minimum size=0.28cm] at (2.3, -2.3) {};
\node[font=\tiny, right] at (2.52, -2.3) {Traversed};
\node[distractor, minimum size=0.28cm] at (3.8, -2.3) {};
\node[font=\tiny, right] at (4.02, -2.3) {Distractor};
\node[font=\tiny, gray] at (6.3, -2.3) {\textcolor{gray!40}{- - -}~Below threshold};

\end{tikzpicture}%
}
\caption{System pipeline. Stage~1 constructs a hybrid document graph (Figure~\ref{fig:graphrag}), selects dense/sparse entry points, retrieves the connected component, and filters disconnected distractors. Stage~2 performs structured analysis-before-answer prompting with self-consistency. Stage~3 applies eight post-hoc consistency heuristics.}
\label{fig:pipeline}
\end{figure*}
\section{System Overview}
\label{sec:system_overview}

Our proposed system follows a three-stage pipeline, illustrated in Figure~\ref{fig:pipeline}: 1)~graph-based retrieval for distractor filtering, 2)~an LLM-based abductive reasoner with reflectively optimized prompt design, 3)~ post-hoc consistency enforcement.
\subsection{Retrieval: Distractor Filtering}
\label{sec:retrieval}

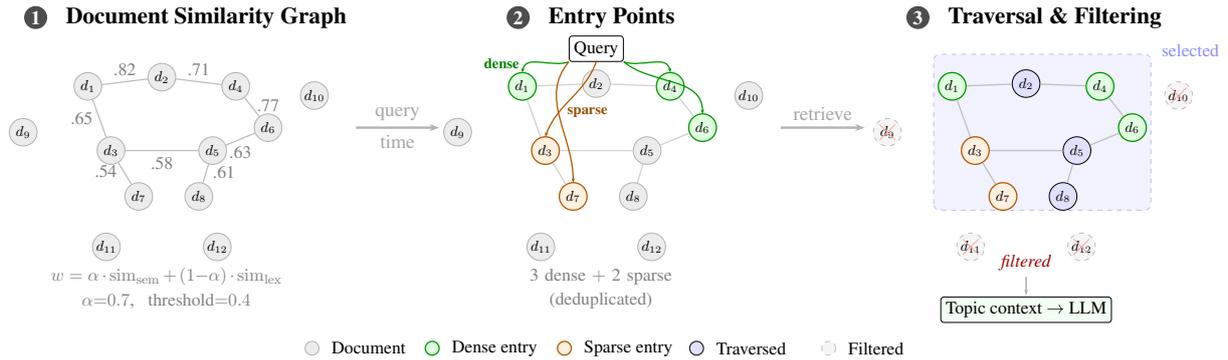
\begin{figure*}[t]
\centering
\resizebox{\textwidth}{!}{%
\begin{tikzpicture}[
    neutral/.style={draw=gray!60, circle, fill=gray!15, minimum size=0.6cm,
                    font=\footnotesize, inner sep=0pt},
    entrydense/.style={draw=green!60!black, circle, fill=green!12, minimum size=0.6cm,
                       font=\footnotesize, inner sep=0pt, line width=0.8pt},
    entrysparse/.style={draw=orange!70!black, circle, fill=orange!12, minimum size=0.6cm,
                        font=\footnotesize, inner sep=0pt, line width=0.8pt},
    doc/.style={draw, circle, fill=blue!12, minimum size=0.6cm,
                font=\footnotesize, inner sep=0pt},
    distractor/.style={draw=gray!50, circle, fill=gray!8, minimum size=0.6cm,
                       font=\footnotesize, inner sep=0pt, dashed},
    edge/.style={gray!50, thick},
    bigarrow/.style={-{Stealth[length=6pt]}, line width=1.5pt, gray!60},
    annot/.style={font=\normalsize, gray},
    badge/.style={fill=black!70, text=white, circle, inner sep=0pt,
                  font=\large\bfseries, minimum size=0.50cm},
    >=Stealth
]

\begin{scope}[xshift=0cm]
    \node[badge] (b1) at (-2.73, 2.40) {1};
    \node[font=\Large\bfseries, anchor=west] at (-2.21, 2.40) {Document Similarity Graph};

    \node[neutral] (p1d1) at (-1.60, 0.90) {$d_1$};
    \node[neutral] (p1d2) at (0.00, 1.10) {$d_2$};
    \node[neutral] (p1d3) at (-1.10, -0.50) {$d_3$};
    \node[neutral] (p1d4) at (1.60, 0.90) {$d_4$};
    \node[neutral] (p1d5) at (1.10, -0.50) {$d_5$};
    \node[neutral] (p1d6) at (2.30, 0.00) {$d_6$};
    \node[neutral] (p1d7) at (-0.50, -1.50) {$d_7$};
    \node[neutral] (p1d8) at (0.80, -1.50) {$d_8$};

    \node[neutral] (p1d9)  at (-3.00, -0.10) {$d_9$};
    \node[neutral] (p1d10) at (3.30, 0.70) {$d_{10}$};
    \node[neutral] (p1d11) at (-1.20, -2.60) {$d_{11}$};
    \node[neutral] (p1d12) at (1.20, -2.60) {$d_{12}$};

    \draw[edge] (p1d1) -- node[above, font=\normalsize, gray] {.82} (p1d2);
    \draw[edge] (p1d1) -- node[left, font=\normalsize, gray] {.65} (p1d3);
    \draw[edge] (p1d2) -- node[above, font=\normalsize, gray] {.71} (p1d4);
    \draw[edge] (p1d3) -- node[below, font=\normalsize, gray] {.58} (p1d5);
    \draw[edge] (p1d4) -- node[right, font=\normalsize, gray, pos=0.4] {.77} (p1d6);
    \draw[edge] (p1d5) -- node[below, font=\normalsize, gray] {.63} (p1d6);
    \draw[edge] (p1d3) -- node[left, font=\normalsize, gray, pos=0.4] {.54} (p1d7);
    \draw[edge] (p1d5) -- node[right, font=\normalsize, gray, pos=0.4] {.61} (p1d8);

    \node[annot, text width=5cm, align=center] at (0.10, -3.25)
        {$w = \alpha \cdot \mathrm{sim_{sem}} + (1{-}\alpha) \cdot \mathrm{sim_{lex}}$};
    \node[annot] at (0.10, -3.77) {$\alpha{=}0.7$, \; threshold${=}0.4$};
\end{scope}

\draw[bigarrow] (4.20, 0) --
    node[above, font=\large, text=gray!70] {query}
    node[below, font=\large, text=gray!70] {time} (6.00, 0);

\begin{scope}[xshift=9.4cm]
    \node[badge] (b2) at (-1.69, 2.40) {2};
    \node[font=\Large\bfseries, anchor=west] at (-1.17, 2.40) {Entry Points};

    \node[draw, rounded corners=2pt, fill=white, font=\normalsize, inner sep=3pt]
        (query) at (0.00, 1.72) {Query};

    \node[entrydense]  (p2d1) at (-1.60, 0.90) {$d_1$};
    \node[neutral]     (p2d2) at (0.00, 0.96) {$d_2$};
    \node[entrysparse] (p2d3) at (-1.10, -0.50) {$d_3$};
    \node[entrydense]  (p2d4) at (1.60, 0.90) {$d_4$};
    \node[neutral]     (p2d5) at (1.10, -0.50) {$d_5$};
    \node[entrydense]  (p2d6) at (2.30, 0.00) {$d_6$};
    \node[entrysparse] (p2d7) at (-0.50, -1.50) {$d_7$};
    \node[neutral]     (p2d8) at (0.80, -1.50) {$d_8$};

    \node[neutral] (p2d9)  at (-3.00, -0.10) {$d_9$};
    \node[neutral] (p2d10) at (3.30, 0.70) {$d_{10}$};
    \node[neutral] (p2d11) at (-1.20, -2.60) {$d_{11}$};
    \node[neutral] (p2d12) at (1.20, -2.60) {$d_{12}$};

    \draw[edge, gray!35] (p2d1) -- (p2d2);
    \draw[edge, gray!35] (p2d1) -- (p2d3);
    \draw[edge, gray!35] (p2d2) -- (p2d4);
    \draw[edge, gray!35] (p2d3) -- (p2d5);
    \draw[edge, gray!35] (p2d4) -- (p2d6);
    \draw[edge, gray!35] (p2d5) -- (p2d6);
    \draw[edge, gray!35] (p2d3) -- (p2d7);
    \draw[edge, gray!35] (p2d5) -- (p2d8);

    \draw[-{Stealth[length=2pt]}, green!60!black, thick]
        (query.south west) to[out=-160, in=80]
        node[left, font=\small\bfseries, green!50!black, pos=0.7, xshift=-3pt] {dense} (p2d1.north);
    \draw[-{Stealth[length=2pt]}, green!60!black, thick]
        (query.south east) to[out=-20, in=100] (p2d4.north);
    \draw[-{Stealth[length=2pt]}, green!60!black, thick]
        (query.south east) to[out=-40, in=90] (p2d6.north);
    \draw[-{Stealth[length=2pt]}, orange!70!black, thick]
        (query.south) to[out=-110, in=70]
        node[right, font=\small\bfseries, orange!60!black, pos=0.7, xshift=1pt] {sparse} (p2d3.north);
    \draw[-{Stealth[length=2pt]}, orange!70!black, thick]
        (query.south west) to[out=-130, in=90] (p2d7.north);

    \node[annot, text width=5cm, align=center] at (0.10, -3.25)
        {3 dense $+$ 2 sparse};
    \node[annot] at (0.10, -3.77) {(deduplicated)};
\end{scope}

\draw[bigarrow] (13.39, 0) --
    node[above, font=\large, text=gray!70] {retrieve} (15.21, 0);

\begin{scope}[xshift=18.7cm]
    \node[badge] (b3) at (-2.34, 2.40) {3};
    \node[font=\Large\bfseries, anchor=west] at (-1.82, 2.40) {Traversal \& Filtering};

    \fill[blue!5, rounded corners=4pt] (-2.00, -1.80) rectangle (2.70, 1.60);
    \draw[blue!25, rounded corners=4pt, dashed] (-2.00, -1.80) rectangle (2.70, 1.60);
    \node[font=\normalsize, blue!45, anchor=west] at (2.80, 1.68) {selected};

    \node[entrydense]  (p3d1) at (-1.60, 0.90) {$d_1$};
    \node[doc]         (p3d2) at (0.00, 0.96) {$d_2$};
    \node[entrysparse] (p3d3) at (-1.10, -0.50) {$d_3$};
    \node[entrydense]  (p3d4) at (1.60, 0.90) {$d_4$};
    \node[doc]         (p3d5) at (1.10, -0.50) {$d_5$};
    \node[entrydense]  (p3d6) at (2.30, 0.00) {$d_6$};
    \node[entrysparse] (p3d7) at (-0.50, -1.50) {$d_7$};
    \node[doc]         (p3d8) at (0.80, -1.50) {$d_8$};

    \node[distractor] (p3d9)  at (-3.00, -0.10) {$d_9$};
    \node[distractor] (p3d10) at (3.30, 0.70) {$d_{10}$};
    \node[distractor] (p3d11) at (-1.20, -2.60) {$d_{11}$};
    \node[distractor] (p3d12) at (1.20, -2.60) {$d_{12}$};

    \draw[edge] (p3d1) -- (p3d2);
    \draw[edge] (p3d1) -- (p3d3);
    \draw[edge] (p3d2) -- (p3d4);
    \draw[edge] (p3d3) -- (p3d5);
    \draw[edge] (p3d4) -- (p3d6);
    \draw[edge] (p3d5) -- (p3d6);
    \draw[edge] (p3d3) -- (p3d7);
    \draw[edge] (p3d5) -- (p3d8);

    \foreach \n in {p3d9, p3d10, p3d11, p3d12} {
        \draw[red!35] ($(\n)+(-0.17,-0.17)$) -- ($(\n)+(0.17,0.17)$);
        \draw[red!35] ($(\n)+(-0.17,0.17)$) -- ($(\n)+(0.17,-0.17)$);
    }
    \node[font=\normalsize\itshape, red!60!black] at (0, -2.93) {filtered};

    \draw[-{Stealth[length=3pt]}, gray!60, line width=1pt]
        (0, -3.25) -- (0, -3.64);
    \node[draw, rounded corners=2pt, fill=green!5, font=\normalsize, inner sep=3pt]
        at (0, -3.97) {Topic context $\rightarrow$ LLM};
\end{scope}

\node[neutral, minimum size=0.3cm] at (3.25, -4.81) {};
\node[font=\normalsize, right] at (3.54, -4.81) {Document};
\node[entrydense, minimum size=0.3cm] at (5.85, -4.81) {};
\node[font=\normalsize, right] at (6.14, -4.81) {Dense entry};
\node[entrysparse, minimum size=0.3cm] at (8.71, -4.81) {};
\node[font=\normalsize, right] at (9.00, -4.81) {Sparse entry};
\node[doc, minimum size=0.3cm] at (11.57, -4.81) {};
\node[font=\normalsize, right] at (11.86, -4.81) {Traversed};
\node[distractor, minimum size=0.3cm] at (14.43, -4.81) {};
\node[font=\normalsize, right] at (14.72, -4.81) {Filtered};

\end{tikzpicture}%
}
\caption{Hybrid document-graph retrieval in three steps. \textbf{Step~1:} Build a hybrid similarity graph ($\alpha{=}0.7$ dense $+$ $0.3$ sparse); disconnected documents ($d_9$--$d_{12}$) are potential distractors. \textbf{Step~2:} At query time, pick entry points from dense and sparse signals (3$+$2, deduplicated). \textbf{Step~3:} Retrieve the full connected component from the seeds, filter disconnected documents, and pass the selected topic context to the LLM reasoner.}
\label{fig:graphrag}
\end{figure*}

\paragraph{Hybrid Document Graphs.}
For each topic, we construct a document similarity graph $G = (V, E)$ (Figure~\ref{fig:graphrag}) where each node $v \in V$ is a context document and edges encode hybrid similarity combining dense and sparse (BM25+ with entity boosting) signals~\cite{lewis2020rag, luan2021sparse}:
\[w(d_i, d_j) = \alpha \cdot \mathrm{sim_{sem}}(d_i, d_j) + (1 - \alpha) \cdot \mathrm{sim_{lex}}(d_i, d_j)\]
$w(d_i, d_j)$ is the hybrid edge weight between documents $d_i$, $d_j$, $\mathrm{sim_{sem}}$ denotes cosine similarity over dense embeddings, $\mathrm{sim_{lex}}$ is BM25+ lexical similarity with entity boosting, and $\alpha{=}0.7$ (Table~\ref{tab:rag_config}).

Unlike entity-centric GraphRAG~\cite{edge2024graphrag}, which extracts named entities as nodes, our graph preserves intact documents, retaining the full narrative context needed for causal reasoning.
At query time, we identify \emph{entry points}, e.g. the top-ranked documents by each signal (3~dense, 2~sparse), deduplicated so that a document retrieved by both signals counts once.
From these seeds, we traverse the graph via breadth-first search (BFS) over the full connected component, collecting every reachable document.
This prioritizes recall over precision: a missing document can break a multi-hop causal chain, whereas extra documents add manageable noise.
The traversal enables multi-hop evidence discovery, where indirectly connected documents may provide critical causal evidence~\cite{wang2025causalrag}.

\begin{table}[t]
\centering
\small
\renewcommand{\arraystretch}{1.15}
\begin{tabular}{@{}lp{4.8cm}}
\hline
\textbf{Component} & \textbf{Description} \\
\hline
Dense embeddings & Cohere Embed v4 (1024-dim) \\
Sparse retrieval & BM25+~\cite{robertson2009bm25} with 3$\times$ entity boosting \\
Hybrid weighting & $\alpha{=}0.7$ dense $+$ $0.3$ sparse \\
Entry points & 3 dense $+$ 2 sparse (deduplicated) \\
Traversal & Full connected component \\
\hline
\end{tabular}
\caption{Hybrid document graph retrieval configuration.}
\label{tab:rag_config}
\end{table}

\paragraph{Topic-Wide Aggregation.}
Since questions within the same topic retrieve heavily overlapping document sets, we aggregate retrieved documents across all questions within each topic into a single \emph{topic-wide context}.
The first question per topic populates the cache, and subsequent questions reuse it~\cite{anthropic2024contextual}, yielding a 91\% cache hit rate and 87\% cost reduction (Table~\ref{tab:rag_stats}).
Retrieval selects 73\% of available documents per topic, excluding 27\% as likely distractors (Table~\ref{tab:rag_results}).

\subsection{Inference}
\label{sec:inference}

\paragraph{Structured Prompting.}
We use an XML-structured prompt with an analysis-before-answer format that forces the LLM to verbalize its reasoning before committing to a final answer.
This acts as a structured scratchpad~\cite{nye2021scratchpads}, eliciting chain-of-thought (CoT) reasoning~\cite{wei2022chainofthought} within a constrained output format: an \texttt{<analysis>} block containing per-option reasoning followed by an \texttt{<answer>} block restricted to valid option letters.
The prompt specifies explicit reasoning criteria (requiring direct textual support and logical sufficiency), selection rules (supporting multi-label answers), and quality checks (enforcing disciplined evaluation per candidate).

\paragraph{Prompt Design.}
As prompt optimizer, we implement GEPA~\cite{agrawal2025gepa} via DSPy~\cite{khattab2024dspy} to explore the space of effective prompt formulations through automatic reflective prompt evolution.
Rather than deploying GEPA-optimized prompts directly as black-box solutions, which raises concerns about data contamination and overfitting to the optimization set, we use the evolved prompts as a source of insight.
The structural heuristics surfaced by GEPA (e.g., single-step causal reasoning, explicit causal language prioritization, duplicate option handling) inform the final prompt design: duplicate option consistency and ``None'' mutual exclusivity are implemented as post-hoc verification heuristics (Section~\ref{sec:postprocessing}) rather than prompt instructions (Appendix~\ref{sec:gepa_insights}).

\paragraph{Self-Consistency.}
We sample $k{=}3$ responses at temperature $\tau{=}1.0$ and aggregate via per-option majority voting~\cite{wang2023selfconsistency, taubenfeld2025confidence}, with an inclusion threshold of 0.5.
For questions containing ``None of the others'', we apply conflict resolution where the higher-voted category prevails.
We compare majority voting against union, intersection, and confidence-weighted aggregation (App.~\ref{sec:aggregation_strategies})
.

\subsection{Post-Hoc Consistency Enforcement}
\label{sec:postprocessing}

We apply eight de-hallucination and consistency-aware heuristics that enforce logical invariants the LLM may violate (e.g. selecting mutually exclusive options simultaneously or assigning different truth values to identical candidates) iteratively until convergence (typically 2~iterations).
Details and per-heuristic impact are reported in Appendix~\ref{sec:verification_rules}.



\section{Experimental Setup}
\label{sec:experimental_setup}

We evaluate our system on the SemEval 2026 Task~12 development set, comparing 18 model configurations under zero-shot structured prompting with optional inference enhancements.

\paragraph{Dataset}
\label{sec:dataset}
The development set comprises 400 questions across 36 topics (775 documents), where zero to four out of the four candidate explanations may be correct.
Notably, 43.6\% of questions have multiple correct answers.
Questions are grouped by topic, with sibling questions sharing document context, a property exploited for caching and consistency enforcement (Sections~\ref{sec:retrieval},~\ref{sec:postprocessing}; App.~\ref{sec:dataset_analysis}).

\paragraph{Evaluation Metric}
\label{sec:evaluation_metric}
We use the task's multi-label classification metric with partial credit.
The final score is computed as the unweighted arithmetic mean across all questions: a model predicts a subset $\hat{y} \subset \{\text{A}, \text{B}, \text{C}, \text{D}\}$ per question, compared against the gold label set $y$ via set-based evaluation:
\begin{itemize}[noitemsep,topsep=0pt]
    \item \textbf{Exact match} ($\hat{y} = y$): score $= 1.0$
    \item \textbf{Partial match} ($\hat{y} \subset y$, $\hat{y} \neq \emptyset$): score $= 0.5$
    \item \textbf{All other cases}: score $= 0.0$
\end{itemize}
Partial credit rewards models that correctly identify at least one valid cause without requiring full recall, thereby providing a finer-grained evaluation signal.

\paragraph{Models}
\label{sec:models}
We evaluate models accessed through AWS Bedrock, Google's Gemini API, and the OpenAI API.
Table~\ref{tab:models} summarizes the evaluated configurations.
\textit{Standard} LLMs use default decoding, while \textit{reasoning} LLMs employ extended thinking (Claude's \texttt{extended\_thinking}, Gemini's \texttt{high} thinking level, OpenAI's \texttt{xhigh} reasoning effort, and DeepSeek R1's native CoT). All models employ zero-shot structured prompting (Section~\ref{sec:inference}), with self-consistency ($k{=}3$, $\tau{=}1.0$)  selectively applied due to cost-related concerns.

\begin{table}[t!]
\centering
\small
\renewcommand{\arraystretch}{1.15}
\begin{tabular}{p{1.1cm}p{2.8cm}p{2.6cm}}
\hline
\textbf{Family} & \textbf{Standard} & \textbf{Reasoning} \\
\hline
Claude    & Haiku 4.5, Sonnet 4,  & Sonnet 4 Thinking,     \\
          & Sonnet 4.5, Opus 4.0, & Sonnet 4.5 Thinking    \\
          & Opus 4.5              &                     \\
Gemini    & ---                   & Flash 3 Preview     \\
OpenAI    & ---                   & GPT-5.2 (xhigh)     \\
DeepSeek  & V3.1                  & V3.1 Think, R1      \\
Llama     & 3.3-70B               & ---                 \\
Kimi      & ---                   & K2 Think            \\
Qwen      & 3-32B                 & ---                 \\
\hline
\end{tabular}
\caption{Model configurations evaluated. 
\emph{Reasoning} models use extended thinking or chain-of-thought.}
\label{tab:models}
\end{table}



\section{Results and Analysis}
\label{sec:results}

Table~\ref{tab:baseline_results} reports zero-shot dev performance across 15 LLMs (scores ranging between 0.611-0.828), highlighting the gap between open-weight and frontier models. Extended thinking consistently helps: Sonnet~4.5 Thinking (0.828) surpasses Sonnet~4.5 by 3.8~pp. GPT-5.2 (0.810) and Gemini~3 Flash (0.804) are competitive, while DeepSeek models underperform relative to their scale and Qwen3-32B (0.611) confirms the mid-size open-weight capacity gap.

\begin{table}[t]
\centering
\small
\renewcommand{\arraystretch}{1.15}
\begin{tabular}{m{1.5cm}lc}
\hline
&\textbf{Model} & \textbf{Score} \\
\hline
\multirow{8}{1.5cm}{\textit{Standard \\ Models}} &
Claude Haiku 4.5       & 0.788 \\
&Claude Sonnet 4        & 0.781 \\
&Claude Sonnet 4.5      & 0.790 \\
&Claude Opus 4.0        & 0.770 \\
&Claude Opus 4.5        & \textbf{0.801} \\
&DeepSeek V3.1          & 0.709 \\
&Llama 3.3-70B          & 0.633 \\
&Qwen3-32B              & 0.611 \\
\hline
\multirow{7}{1.5cm}{\textit{Reasoning \\ Models}} &
Claude Sonnet 4 Thinking  & 0.794 \\
&Claude Sonnet 4.5 Thinking & \textbf{0.828} \\
&Gemini 3 Flash Preview & 0.804 \\
&GPT-5.2                & 0.810 \\
&DeepSeek V3.1 Think    & 0.774 \\
&DeepSeek R1            & 0.718 \\
&Kimi K2 Think          & 0.776 \\
\hline
\end{tabular}
\caption{Zero-shot baseline results (dev set). \textbf{Bold} marks the best score within each group under the partial-credit metric (Section~\ref{sec:evaluation_metric}).}
\label{tab:baseline_results}
\end{table}



Graph-based retrieval (Section~\ref{sec:retrieval}) provides marginal accuracy gains for frontier LLMs but yields a $+$9~pp improvement for Haiku~3.5 (Table~\ref{tab:rag_results}), consistent with findings that retrieval disproportionately benefits smaller models~\cite{xu2024retrieval}, while reducing cost by 87\% via topic-wide caching, mitigating accuracy degradation~\cite{liu2024lost, du2025context} and inference cost~\cite{li2024rag_vs_lc} of long-context processing.
Self-consistency ($k{=}3$, $\tau{=}1.0$) yields modest gains (e.g., $+$1.6~pp for Sonnet~4.5); majority voting across three systems reaches 0.811 (Table~\ref{tab:sc_results}, App.~\ref{sec:aggregation_strategies}).
Post-hoc heuristics (Section~\ref{sec:postprocessing}) produces the single largest gain: Sonnet~4.5 Thinking improves from 0.828 to 0.884 ($+$5.6~pp), with 85.4\% of corrections being genuine improvements. Cross-question checks contribute the largest individual gains (details in App.~\ref{sec:verification_rules}).

\paragraph{Test Set}
Table~\ref{tab:test_results} reports results on the held-out test set (612 questions, 18.3\% multi-answer).
Post-hoc heuristics again yield the largest gains ($+$4.8~pp for Claude, $+$3.7~pp for GPT-5.2), consistent with the dev-set pattern, while self-consistency alone has minimal impact: the gain materializes only when combined with post-hoc heuristics (0.943, $+$4.1~pp).
The multi-model ensemble reaches 0.926 without heuristics but falls short of the best individual model with heuristics (Claude Sonnet 4.5 Thinking at 0.952).
The higher test-set scores reflect the lower multi-answer prevalence (18.3\% vs.\ 47.5\%), consistent with the inherently greater difficulty of multi-answer questions.

\begin{table}[t!]
\centering
\small
\begin{tabular}{p{1.2cm}lp{0.4cm}r}
\hline
&\textbf{Configuration} & \textbf{Base} & \textbf{+ Post-hoc} \\
\hline
\multirow{3}{1cm}{\textit{Individual \\ Models}}
&Claude Son.~4.5 Thinking & 0.904 & \textbf{0.952} \\
&GPT-5.2                   & 0.912 & 0.949 \\
&Gemini 3 Flash             & 0.907 & 0.943 \\
\hline
\multirow{2}{1.2cm}{\textit{Self-Consistency}} 
&SC: Sonnet 3$\times$ ($\theta{=}0.50$)  & 0.902 & 0.948 \\
&SC: Gemini 5$\times$ ($\theta{=}0.50$)  & 0.902 & 0.943 \\
\hline
\textit{Ensemble} & Sonnet + GPT + Gemini      & 0.926 & 0.952 \\
\hline
\end{tabular}
\caption{Test set results. Post-hoc = deterministic consistency enforcement via heuristics. SC = self-consistency with majority voting; $\theta$ = inclusion threshold.}
\label{tab:test_results}
\end{table}

\subsection{Error Analysis}
\label{sec:error_analysis}

LLMs exhibit systematic biases in causal inference, including the \emph{post-hoc} fallacy~\cite{joshi2024llms} and degraded multi-hop reasoning~\cite{chi2024unveiling, yu2025causaleval}.
Cross-validation across all 14 models (7~families) yields Fleiss'~$\kappa = 0.690$~\cite{fleiss1971measuring} (\emph{substantial} agreement), with within-family agreement ($\kappa = 0.794$) exceeding cross-family agreement ($\kappa = 0.661$; Figure~\ref{fig:agreement_heatmap}).
Only 3 questions are truly \emph{unbreakable} (all 14 models score zero), but 42 receive no exact match from any model, 38 of which are multi-answer under-selection failures (Figure~\ref{fig:failure_gradient}).
An oracle selecting the best model per question achieves 0.895, indicating 6.7~pp headroom from model complementarity (App.~\ref{sec:ensemble_appendix}).

\paragraph{Single vs.\ Multi-Answer Performance.}
All models show lower multi-answer accuracy (Table~\ref{tab:single_multi_gap}), with gaps of 25-56~pp across 14 models. Total under-selections (1,389) vastly exceed over-selections (52), confirming conservative cause selection as the dominant bias (Figure~\ref{fig:model_accuracy}).

\begin{table}[t]
\centering
\small
\renewcommand{\arraystretch}{1.15}
\begin{tabular}{@{}lrrr@{}}
\hline
\textbf{System} & \textbf{Single} & \textbf{Multi} & $\boldsymbol{\Delta}$ \\
\hline
GPT-5.2              & 91.0 & 52.6 & $-$38.4 \\
Claude Son.~4.5 Thinking & 88.6 & 60.5 & $-$28.1 \\
Gemini 3 Flash        & 85.2 & 63.7 & $-$21.5 \\
\hline
\end{tabular}
\caption{Exact-match accuracy (\%) by answer cardinality across three frontier systems. Multi-answer questions (190/400, 47.5\%) are inherently harder due to the combinatorial requirement.}
\label{tab:single_multi_gap}
\end{table}

\paragraph{Failure Taxonomy.}
Of the 42 no-exact-match questions (3~truly unbreakable; Table~\ref{tab:unbreakable}, App.~\ref{sec:failure_taxonomy}), models select on average 1.2 options when gold requires 2.4, a 51\% cause-count reduction, and in 83\% of cases the consensus is a \emph{single option} (App.~\ref{sec:mechanism_analysis}).
Three shared inductive biases persist across all seven model families:
(1)~\emph{causal chain incompleteness}~\cite{chi2024unveiling, yu2025causaleval}, the most prevalent pattern (18/42), where models select one link of a multi-step causal chain and omit the rest;
(2)~\emph{proximate cause preference}~\cite{joshi2024llms} (11/42), which favors the most recent antecedent over enabling conditions; and
(3)~\emph{salience bias} (9/42), where models select dramatic, newsworthy causes over subtler contributing factors.
These biases are correlated across families (error overlap ratio $2.0\times$ above independence), suggesting shared pretraining priors rather than prompt-specific artifacts~\cite{chi2024unveiling}.


\section{Conclusion}

Our proposed three-stage system (graph-based retrieval, GEPA-informed prompt design, deterministic consistency enforcement via heuristics) ranks first on the evaluation-phase leaderboard (0.95), with post-hoc heuristics providing the largest dev-set gain ($+$5.6~pp).
Our error analysis across 14 models (7~families) identifies three shared inductive biases: causal chain incompleteness (18/42), proximate cause preference (11/42), and salience bias (9/42), which together produce a single-cause default (51\% cause-count reduction, 83\% of failures converging on one option).
With an oracle upper bound of 0.895 (best of 14 models per question), no single family dominates across all question types, suggesting that targeted ensemble or multi-agent architectures could close much of the remaining gap.

\section*{Limitations}
\paragraph{Task-Specific Assumptions.}
The post-hoc consistency enforcement via heuristics step exploits task-specific structural properties (topic groupings, duplicate options, ``None'' exclusivity) and would require adaptation for other benchmarks.

\paragraph{Model Coverage.}
Our evaluation focuses on frontier-scale to mid-size (32B) models; smaller open-weight models remain unexplored.


\bibliography{custom}

\appendix


\section{Exploratory Data Analysis}
\label{sec:dataset_analysis}

\subsection{Dataset formulation}
The questions file is represented in a jsonl object with the following format for a specific instance:

\begin{tcolorbox}[
  colback=palepurple,
  colframe=black!30,
  boxrule=0.5pt,
  arc=2mm,
  left=6pt,
  right=6pt,
  top=6pt,
  bottom=3pt
]
\begin{lstlisting}[language=json]
{
  "topic_id": 11,
  "id": "q-101",
  "target_event": "Short description of an observed event.",
  "option_A": "Candidate explanation A.",
  "option_B": "Candidate explanation B.",
  "option_C": "Candidate explanation C.",
  "option_D": "Candidate explanation D.",
  "golden_answer": "A,B"
}
\end{lstlisting}
\end{tcolorbox}

\noindent\textbf{Example.} Input instance object. \newline

Consequently, the context comprises  records in the docs file, with a jsonl instance as following:

\begin{tcolorbox}[
  colback=palepurple,
  colframe=black!30,
  boxrule=0.5pt,
  arc=2mm,
  left=6pt,
  right=6pt,
  top=6pt,
  bottom=3pt
]
\begin{lstlisting}[language=json]
{
  "topic_id": 11,
  "topic": "OpenAI releases ChatGPT.",
  "docs": [
    {
      "title": "Article title",
      "id": "doc-001",
      "link": "https://example.com",
      "snippet": "Short summary of the document.",
      "source": "News source",
      "imageUrl": "Base64imageUrl",
      "content": "Full document text."
    }
  ]
}
\end{lstlisting}
\end{tcolorbox}
\noindent\textbf{Example.} Background context for an event. \newline

\subsection{Overview and Statistics}

The dataset comprises 2,831 questions distributed across multiple splits as shown in Table~\ref{tab:dataset_overview}.
Each question presents a target event and four candidate explanations, where one or more may be correct causes.

\begin{table}[!ht]
\centering
\small
\begin{tabular}{lrrr}
\toprule
\textbf{Statistic} & \textbf{Train} & \textbf{Dev} & \textbf{Test} \\
\midrule
Questions & 1,819 & 400 & 612 \\
Topics & 36 & 36 & 24 \\
Documents & 775 & 775 & 405 \\
Multi-answer (\%) & 42.7 & 47.5 & -- \\
Avg. Q/Topic & 50.5 & 11.1 & 25.5 \\
Avg. Doc Length & 1188 & 1188 & 899 \\
\bottomrule
\end{tabular}
\caption{Dataset statistics across splits. Multi-answer percentage indicates questions with more than one correct answer.}
\label{tab:dataset_overview}
\end{table}

\begin{table*}[!t]
\centering
\small
\renewcommand{\arraystretch}{1.15}
\begin{tabular}{@{}p{4.5cm}p{9cm}c@{}}
\toprule
\textbf{Target Event} & \textbf{Candidate Options} & \textbf{Gold} \\
\midrule
Ontario placed a 25\% tax on electricity exports to the U.S.
& \textbf{(A) Trump placed 25\% tariffs on imports from Canada.} \newline
  (B) McKinley was shot by anarchist Czolgosz in 1901. \newline
  (C) The 16th Amendment was ratified in 1913. \newline
  (D) McKinley died on Sept.\ 14, 1901.
& A \\
\midrule
The Chinese government criticized Japan's plan to release Fukushima wastewater.
& (A) IAEA confirmed the release would have negligible radiological effect. \newline
  \textbf{(B) Japan's cabinet approved a treated-water plan in 2021.} \newline
  \textbf{(C) Japan announced its plan to release treated wastewater into the sea.} \newline
  (D) TEPCO has used water to cool nuclear fuel rods since 2011.
& B,C \\
\midrule
Facebook changed its corporate name to Meta at the annual Connect event.
& \textbf{(A) Zuckerberg introduced Meta at the Connect 2021 event.} \newline
  (B) Facebook announced Bosworth's promotion to CTO. \newline
  \textbf{(C) Facebook plans to change its name to reflect metaverse focus.} \newline
  \textbf{(D) Zuckerberg announced Facebook was betting on the metaverse.}
& A,C,D \\
\midrule
Goldman Sachs estimated AI could automate 300 million jobs worldwide.
& (A) OpenAI released ChatGPT Plus and later introduced GPT-4 Turbo. \newline
  (B) Stack Overflow banned ChatGPT-generated answers. \newline
  \textbf{(C) None of the others are correct causes.} \newline
  (D) OpenAI launched DALL-E 2 beta in July 2022.
& C \\
\bottomrule
\end{tabular}
\caption{Representative examples illustrating each answer cardinality. Correct options are shown in \textbf{bold}. Option texts are abbreviated for space.}
\label{tab:examples}
\end{table*}

\begin{figure}[!t]
\centering
\includegraphics[width=\columnwidth]{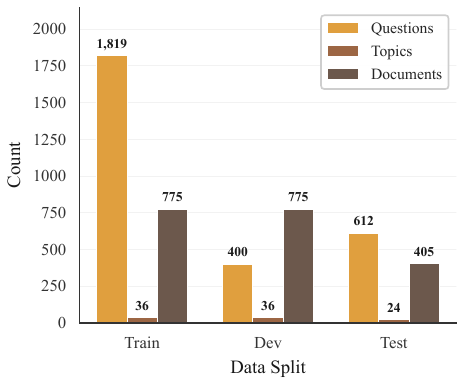}
\caption{Dataset composition across splits.}
\label{fig:composition}
\end{figure}

Figure~\ref{fig:composition} illustrates the composition of each data split, showing the distribution of questions, topics, and documents.

Table~\ref{tab:examples} shows representative examples from the dataset, one for each answer cardinality.
Each question presents a target event and four candidate options; the task is to identify all plausible causes.

\subsection{Answer Distribution}

\begin{figure}[!t]
\centering
\includegraphics[width=0.95\columnwidth]{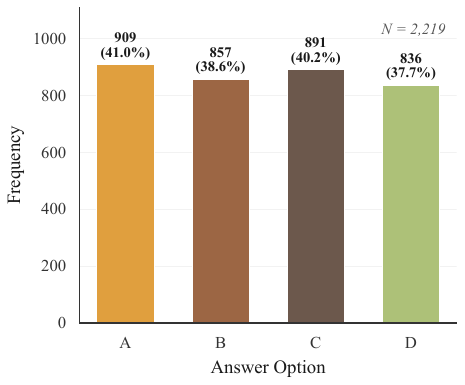}
\caption{Answer frequency by option.}
\label{fig:answer_freq}
\end{figure}

\begin{figure}[!t]
\centering
\includegraphics[width=0.85\columnwidth]{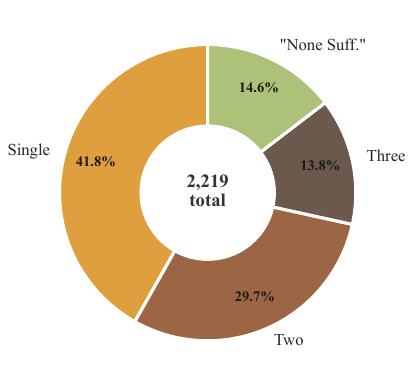}
\caption{Distribution of answer cardinality.}
\label{fig:cardinality}
\end{figure}

\begin{figure}[!t]
\centering
\includegraphics[width=\columnwidth]{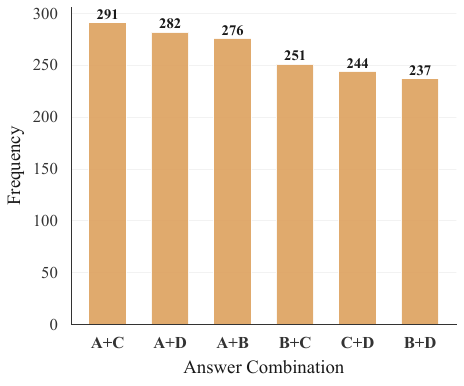}
\caption{Most common multi-answer combinations.}
\label{fig:answer_combos}
\end{figure}

A distinctive characteristic of this dataset is its multi-label nature: 43.6\% of questions have multiple correct answers. Table~\ref{tab:answer_dist} details the answer distribution.
\begin{table}[!t]
\centering
\small
\begin{tabular}{lrr}
\toprule
\textbf{Category} & \textbf{Count} & \textbf{Percentage} \\
\midrule
Option A selected & 909 & 41.0\% \\
Option B selected & 857 & 38.6\% \\
Option C selected & 891 & 40.2\% \\
Option D selected & 836 & 37.7\% \\
\midrule
Single answer & 928 & 41.8\% \\
Two answers & 660 & 29.7\% \\
Three answers & 307 & 13.8\% \\
``None sufficient'' & 324 & 14.6\% \\
\bottomrule
\end{tabular}
\caption{Answer distribution analysis on the combined train and dev sets.}
\label{tab:answer_dist}
\end{table}
The answer options (A, B, C, D) are roughly balanced, indicating no positional bias in the correct answers.
Figure~\ref{fig:answer_freq} shows the frequency distribution, and Figure~\ref{fig:cardinality} visualizes the answer cardinality breakdown.
Figure~\ref{fig:answer_combos} illustrates the most common multi-answer combinations, revealing patterns in how answers co-occur.

\subsection{Multi-Label Characteristics}

The multi-label nature of the dataset poses unique challenges for evaluation. We compute standard multi-label metrics, such as label cardinality, density and multi-label rate (Table \ref{tab:multilabel}).

\begin{table}[!t]
\centering
\small
\begin{tabular}{lr}
\toprule
\textbf{Metric} & \textbf{Value} \\
\midrule
Label cardinality & 1.574 \\
Label density & 0.394 \\
Multi-label rate & 43.6\% \\
\bottomrule
\end{tabular}
\caption{Multi-label characteristics of the dataset.}
\label{tab:multilabel}
\end{table}

The label cardinality of 1.574 indicates that on average, each question has approximately 1.5 correct answers.
This is higher than typical single-label QA datasets and reflects the inherent ambiguity in causal reasoning, since multiple events can plausibly cause a given outcome.

\begin{figure}[!t]
\centering
\includegraphics[width=0.85\columnwidth]{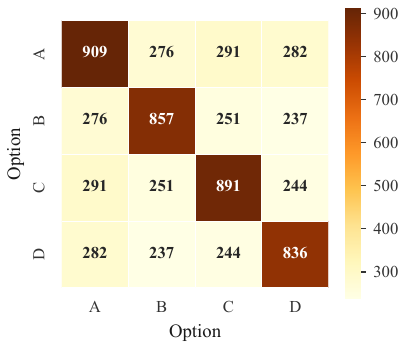}
\caption{Answer co-occurrence matrix.}
\label{fig:cooccurrence}
\end{figure}

Figure~\ref{fig:cooccurrence} shows the co-occurrence matrix of answer labels, revealing which answer pairs frequently appear together in multi-label questions. There are no significant discrepancies in label combination frequencies; combinations containing label A are more prevalent, since A is the most frequent answer option in the dataset (Figure \ref{fig:answer_freq}).

\subsection{Topic Analysis}

The dataset covers 36 real-world news topics spanning significant events from 2016 to 2025, including political events (Brexit, US elections), technological developments (ChatGPT, Huawei), and natural disasters (California wildfires, earthquakes).

\begin{table*}[!t]
\centering
\small
\begin{tabular}{clrrrr}
\toprule
\textbf{ID} & \textbf{Topic Description} & \textbf{Questions} & \textbf{Docs} & \textbf{Multi\%} & \textbf{Context} \\
\midrule
19 & Historic winter storm paralyzes Texas, causin... & 132 & 24 & 53.0 & 18,289 \\
30 & Killing of George Floyd by police sparks Blac... & 113 & 25 & 57.5 & 70,271 \\
23 & Japan begins discharging treated Fukushima nu... & 111 & 24 & 45.0 & 38,702 \\
24 & Halloween crowd crush in Seoul's Itaewon kill... & 111 & 28 & 45.0 & 14,375 \\
22 & Iranian General Qasem Soleimani killed by US ... & 110 & 20 & 36.4 & 12,160 \\
32 & Taliban regains control of Afghanistan after ... & 109 & 27 & 55.0 & 24,098 \\
21 & Myanmar military stages coup, detains Aung Sa... & 106 & 29 & 70.8 & 27,639 \\
3 & Trump supporters storm US Capitol & 91 & 19 & 57.1 & 15,253 \\
25 & Yellow Vest movement sparks widespread protes... & 77 & 18 & 51.9 & 17,805 \\
13 & Facebook changes company name to Meta to focu... & 76 & 21 & 46.1 & 26,242 \\
\bottomrule
\end{tabular}
\caption{Topic-level statistics (top 10 topics by question count). Full table in supplementary materials.}
\label{tab:topic_analysis}
\end{table*}

\begin{figure*}[!t]
\centering
\includegraphics[width=0.95\textwidth]{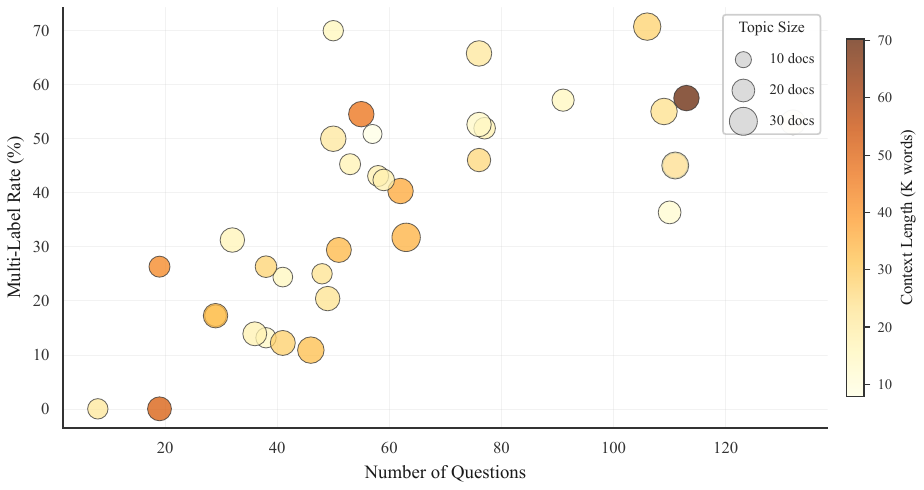}
\caption{Topic characteristics: questions vs complexity (bubble size = document count, color = context length).}
\label{fig:topic_scatter}
\end{figure*}

Figure~\ref{fig:topic_scatter} provides a comprehensive view of topic characteristics, showing the relationship between question count, complexity, and context size.

\subsection{Document Corpus}

\begin{figure}[!t]
\centering
\includegraphics[width=0.97\columnwidth]{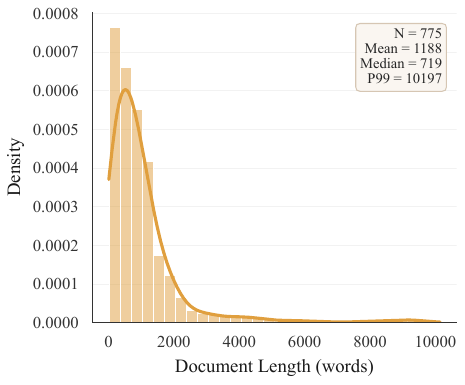}
\caption{Distribution of document lengths (99th percentile shown).}
\label{fig:doc_length}
\end{figure}

Each topic is associated with a set of retrieved documents providing context for causal reasoning.
Figure~\ref{fig:doc_length} shows the distribution of document lengths.

Document lengths vary substantially, from 11 to 44,818 words (median: 719 words), reflecting the heterogeneous, long-tailed nature of web-sourced content.
This variation presents challenges for models with limited context windows. Moreover, the long-tailed nature of this curve may trigger length biases in LLMs, showing preference to the most frequently occurring length over truly consulting the documents' context.

\subsection{Linguistic Analysis}

\begin{figure}[!t]
\centering
\includegraphics[width=0.95\columnwidth]{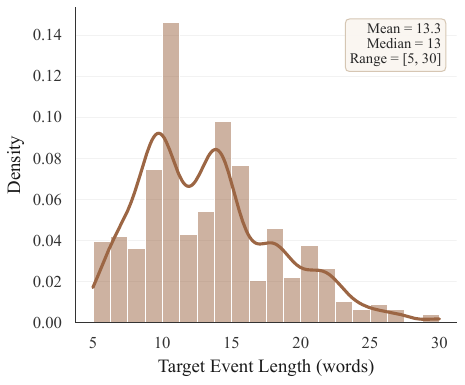}
\caption{Target event length distribution.}
\label{fig:event_length}
\end{figure}

Target events are concise descriptions averaging 13.3 words (Figure~\ref{fig:event_length}).
Candidate options have similar lengths across positions (A through D), ranging from 12.6 to 12.8 words on average. This uniform length distribution in response candidates relieve possible length biases in response selection.

\subsection{Logical Consistency Properties}

We identify two logical consistency properties inherent in the question structure that LLMs may occasionally violate in their predictions.

\paragraph{Mutual Exclusivity of ``None of the Others.''}
14.6\% of questions include a ``None of the others are correct causes'' option.
By definition, this option is mutually exclusive with all other candidates and selecting it alongside other options constitutes a logical contradiction.
However, LLMs are not constrained to respect this exclusivity, and may produce predictions that include ``None'' together with other candidates.

\paragraph{Duplicate Option Consistency.}
22.9\% of questions contain duplicate answer options (identical text appearing in two or more positions), an artifact of the dataset construction process.
Identical options necessarily share the same truth value and thus selecting one but not the other would be logically inconsistent.
These cases expose instances where LLM predictions lack internal consistency.

These observations motivate a set of post-hoc consistency heuristics applied to verify or correct model predictions (Section~\ref{sec:verification_rules}).

\subsection{Post-Hoc Prediction Refinement}
\label{sec:verification_rules}

We apply a set of deterministic post-hoc consistency heuristics that enforce logical invariants the LLM may violate during inference.
These heuristics operate entirely on the model's own predictions acting as restorative heuristics, requiring no gold labels or additional LLM calls.

\begin{figure}[!t]
\centering
\includegraphics[width=\columnwidth]{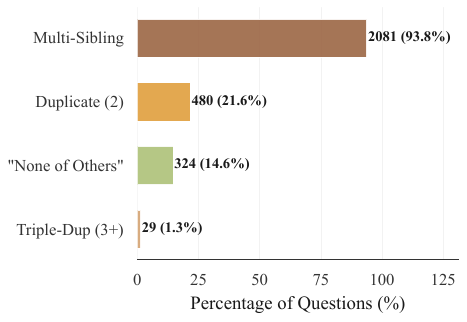}
\caption{Frequency of structural patterns addressed by post-hoc consistency via heuristics.}
\label{fig:pattern_freq}
\end{figure}

\begin{figure}[!t]
\centering
\includegraphics[width=0.9\columnwidth]{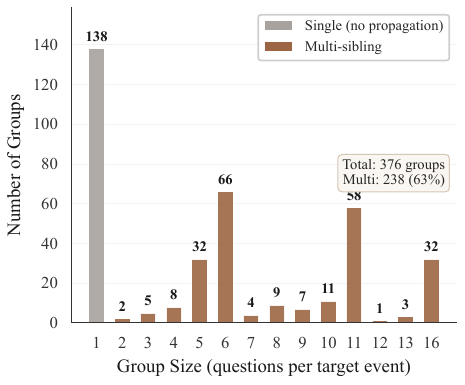}
\caption{Sibling group size distribution. Groups of 2+ questions enable cross-question consistency enforcement.}
\label{fig:sibling_dist}
\end{figure}

The core heuristics address two types of inconsistency (Figure~\ref{fig:pattern_freq}):
(1)~mutual exclusivity violations, where ``None of the others'' is predicted alongside other options (14.6\% of questions);
and (2)~duplicate inconsistencies, where identical options receive different predictions (21.6\% of questions).
Additional heuristics enforce structural invariants across sibling questions e.g. questions sharing the same target event (63.3\% of sibling groups contain 2+ questions; Figure~\ref{fig:sibling_dist}).
Since identical option text must carry the same truth value regardless of which question it appears in, and since ``None of the others'' logically excludes its co-occurring candidates across all siblings, these invariants propagate the logical consequences of the model's within-question judgments to related questions.

Table~\ref{tab:verification_rules} describes all eight heuristics explored.

\paragraph{Iterative Application.}
Heuristics from Table~\ref{tab:verification_rules}   are applied iteratively until convergence (fixed-point), as resolving one inconsistency may reveal another.
For example, enforcing ``None'' exclusivity (R1) triggers cross-question exclusion (R6), which may leave a single valid option, triggering propagation (R7) or closure (R8).
Similarly, enforcing duplicate consistency (R2) triggers cross-sibling propagation (R4), which may trigger R2 again in receiving questions.
Convergence typically requires 2 iterations.

\begin{figure*}[!t]
\centering
\includegraphics[width=\textwidth]{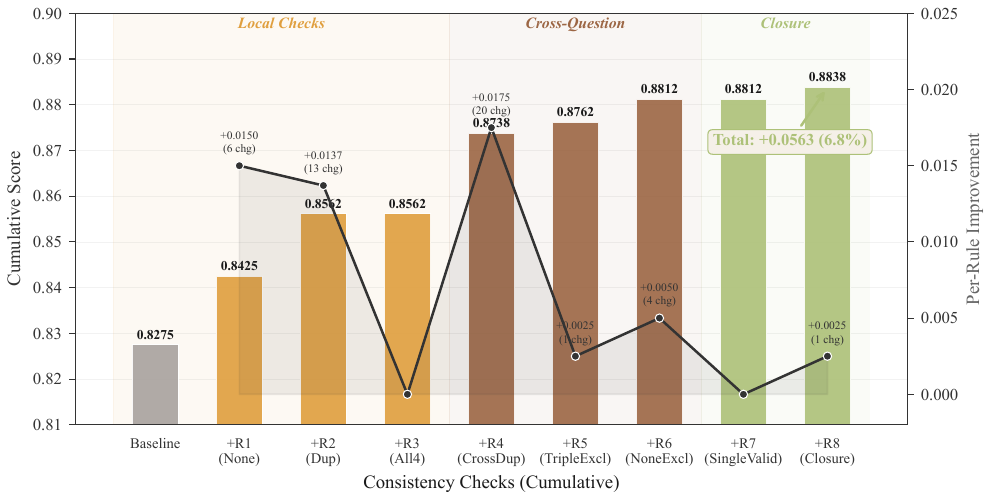}
\caption{Cumulative accuracy improvement from post-hoc consistency heuristics on Claude Sonnet 4.5 (thinking) predictions. Background colors indicate heuristic scope: local (blue), cross-question (red), and closure (green). Black line shows per-rule improvement.}
\label{fig:score_impact}
\end{figure*}

Figure~\ref{fig:score_impact} shows the cumulative score improvement as heuristics are added incrementally.
Cross-question consistency heuristics (R4, R6) provide the largest gains, confirming that enforcing structural invariants across siblings is the most effective signal.
On the dev set, these consistency heuristics improve accuracy from 0.8275 to 0.8838 (+6.8\%), with 85.4\% of corrections being improvements and 14.6\% neutral (changing one incorrect prediction to another also incorrect response), with zero cases where a correct prediction was made incorrect.
The absence of degradations indicates that the structural invariants enforced by these heuristics (duplicate truth-value consistency and mutual exclusivity of ``None'') are well-aligned with the ground truth.

\begin{table*}[!h]
\centering
\small
\begin{tabular}{clllr}
\toprule
\textbf{Heuristic} & \textbf{Name} & \textbf{Scope} & \textbf{Mechanism} & \textbf{Changes} \\
\midrule
R1 & None Exclusivity & Local & Enforce mutual exclusivity of ``None'' option & 6 \\
R2 & Duplicate Consistency & Local & Ensure identical options receive same prediction & 13 \\
R3 & Over-selection Guard & Local & Resolve over-selection when all four options predicted & 0 \\
R4 & Cross-Q Dup.\ Prop. & Cross-Q & Propagate duplicate consistency to sibling questions & 20 \\
R5 & Triple Exclusion & Cross-Q & Exclude unique option from triple-duplicate siblings & 1 \\
R6 & None Cross-Exclusion & Cross-Q & Extend None exclusivity to sibling questions & 4 \\
R7 & Single-Valid Prop. & Cross-Q & Propagate last valid option to siblings & 0 \\
R8 & Closure & Local & Select only remaining valid option & 1 \\
\bottomrule
\end{tabular}
\caption{Post-hoc consistency heuristics applied iteratively until convergence (fixed-point). Scope indicates whether the heuristic operates within a single question (Local) or across sibling questions sharing the same target event (Cross-Q). Changes reported on Claude Sonnet 4.5 (thinking), dev set.}
\label{tab:verification_rules}
\end{table*}

\section{Graph-Based Retrieval}
\label{sec:rag_appendix}

This appendix formalizes the retrieval algorithm and provides visualization examples of the hybrid document graph retrieval system described in Section~\ref{sec:retrieval}.

\subsection{Retrieval Algorithm}
\label{sec:retrieval_algorithm}

Algorithm~\ref{alg:graph_retrieval} formalizes the hybrid graph retrieval procedure described in Section~\ref{sec:retrieval}.

\begin{algorithm}[t] 
\caption{Hybrid Graph Retrieval}
\label{alg:graph_retrieval}
\begin{algorithmic}[1]
\Require Document graph $G = (V, E)$ with hybrid edge weights $w$; query $q$; dense embedder $\phi$; BM25+ index $\mathcal{B}$; parameters $k_{\mathrm{sem}}, k_{\mathrm{lex}}, \tau$
\Ensure Selected document set $\mathcal{S} \subseteq V$

\Statex \textbf{--- Entry Point Identification ---}
\State $\mathcal{E}_{\mathrm{sem}} \gets \text{top-}k_{\mathrm{sem}} \text{ docs by } \cos(\phi(q), \phi(d))$
\State $\mathcal{E}_{\mathrm{lex}} \gets \text{top-}k_{\mathrm{lex}} \text{ docs by } \mathcal{B}(q)$
\State $\mathcal{E} \gets \mathcal{E}_{\mathrm{sem}} \cup \mathcal{E}_{\mathrm{lex}}$ \Comment{Deduplicate}

\Statex \textbf{--- Connected-Component Traversal (BFS) ---}
\State $\mathcal{S} \gets \mathcal{E}$ \Comment{Initialize with entry points}
\State $\textsc{Queue} \gets \mathcal{E}$
\While{$\textsc{Queue} \neq \emptyset$}
    \State $d \gets \textsc{Queue}.\text{dequeue}()$
    \For{each neighbor $d'$ s.t.\ $(d, d') \in E$ and $w(d, d') \geq \tau$}
        \If{$d' \notin \mathcal{S}$}
            \State $\mathcal{S} \gets \mathcal{S} \cup \{d'\}$
            \State $\textsc{Queue}.\text{enqueue}(d')$
        \EndIf
    \EndFor
\EndWhile

\Statex \textbf{--- Distractor Filtering ---}
\State \Return $\mathcal{S}$ \Comment{Disconnected docs excluded}
\end{algorithmic}
\end{algorithm}

The algorithm prioritizes \textit{recall over precision}: a missing document can break a multi-hop causal chain, whereas extra documents add manageable noise.
Entry points are identified via both semantic (dense embedding cosine similarity) and lexical (BM25+ with $3\times$ entity boosting) signals, then deduplicated.
The BFS traversal expands from these seeds through the full connected component, ensuring that indirectly related documents providing bridging causal evidence are included.

\subsection{Document Graph and Query-Time Selection}
\label{sec:doc_context_graph}

\begin{figure*}[!h]
\centering
\includegraphics[width=0.86\textwidth]{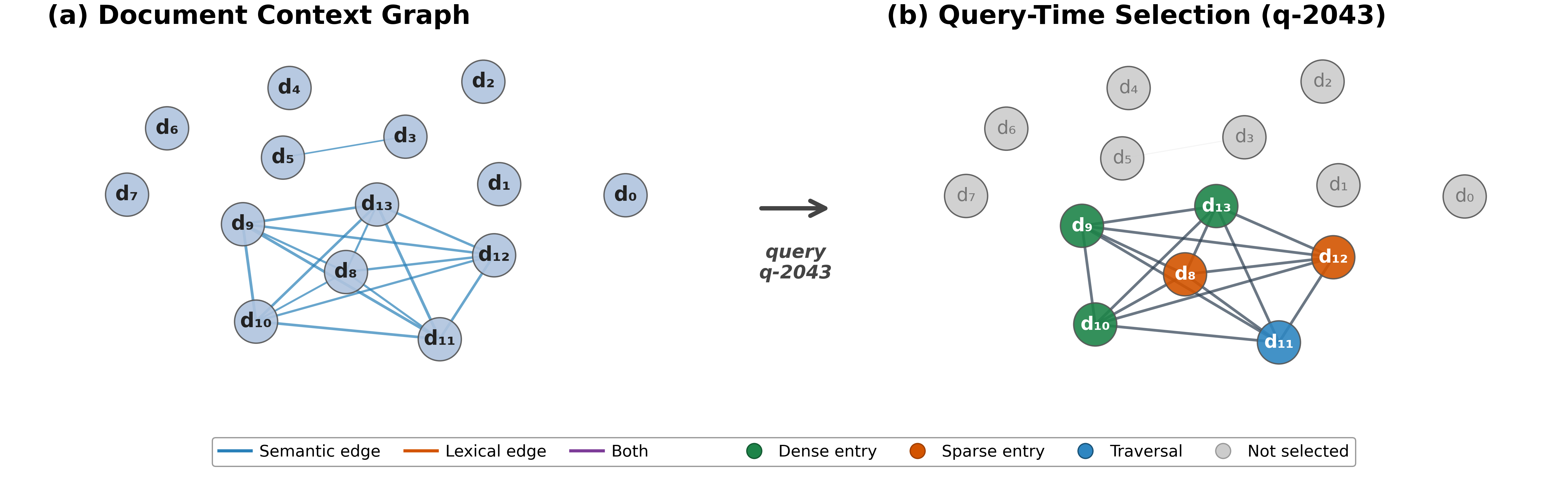}
\caption{Document graph and query-time selection for Topic~7 (South Korean impeachment crisis). Node positions are identical across panels. \textbf{(a)} Document context graph: 14 documents connected by hybrid similarity edges (semantic-dominant, lexical-dominant, balanced). The densely connected component captures the core impeachment narrative; peripheral nodes represent tangentially related events. \textbf{(b)} Query-time selection for q-2043 (target: ``Yoon's aides offered to resign''): 6/14 documents (43\%) selected via 3 dense entries, 2 sparse entries, and 1 traversal-discovered node. Grey nodes are filtered distractors.}
\label{fig:graph_rag_combined}
\end{figure*}

For each topic, we construct a document similarity graph where nodes represent documents and edges encode hybrid similarity scores (weighted combination of dense and sparse signals).
Figure~\ref{fig:graph_rag_combined} illustrates the complete retrieval process for Topic~7 (``South Korean President Yoon Suk-yeol faces impeachment crisis''), which contains 14 documents across 8 connected components.

Panel~(a) shows the full graph structure with edges colored by their dominant similarity signal.
At query time~(b), dense and sparse retrieval identify entry points, then BFS traversal expands through the connected component.
The retrieval selects 6 of 14 documents (43\%) through three mechanisms: 3 semantic entry points (dense embedding matches), 2 lexical entry points (BM25+ matches), and 1 traversal-discovered document (reached via graph edges).
The remaining 8 documents are excluded as likely distractors: they discuss related but distinct events that would add noise without providing direct causal evidence.
This example demonstrates how graph traversal enables multi-hop evidence discovery: the traversal node (document~11) was not a direct match for the query but is connected to selected documents via the similarity graph, providing additional context that supports causal reasoning.
Across the full dev set, retrieval selects 73\% of documents on average, with a 91\% topic-wide cache hit rate (Section~\ref{sec:retrieval}).

%



\begin{table}[t]
\centering

\small
\renewcommand{\arraystretch}{1.15}
\begin{tabular}{@{}lr@{}}
\toprule
\textbf{Metric} & \textbf{Value} \\
\midrule
Avg.\ documents per topic & 21.5 \\
Avg.\ retrieved per question & 16.6 (73\%) \\
Avg.\ in topic-wide context & 15.8 (73\%) \\
Likely distractors excluded & 27\% \\
Prompt cache hit rate & 91\% \\
Overall inference cost reduction & 87\% \\
\bottomrule
\end{tabular}
\caption{Retrieval statistics on the dev set (400 questions, 36 topics, 775 documents).}
\label{tab:rag_stats}
\end{table}

\begin{table}[t]
\centering
\small
\begin{tabular}{@{}lccr@{}}
\toprule
\textbf{Model} & \textbf{Base} & \textbf{+Retr.} & $\boldsymbol{\Delta}$ \\
\midrule
Haiku 3.5\textsuperscript{\dag} & 0.440 & 0.530 & $+$0.090 \\
Sonnet 4              & 0.781 & 0.785 & $+$0.004 \\
Gemini 3 Flash        & 0.804 & 0.806 & $+$0.003 \\
GPT-5.2               & 0.810 & 0.818 & $+$0.008 \\
\bottomrule
\multicolumn{4}{@{}l@{}}{\textsuperscript{\dag}\footnotesize 100 questions; others on full dev set (400).}
\end{tabular}
\caption{Impact of distractor filtering on accuracy. Smaller models benefit substantially from context reduction~\cite{xu2024retrieval}.}
\label{tab:rag_results}
\end{table}


\section{Prompt Templates}
\label{sec:prompts}

In this section, we present the key prompt templates used in our system.
All prompts leverage XML-structured formatting to enforce disciplined reasoning and constrained output (Section~\ref{sec:inference}).
Placeholder tokens in brackets are filled at inference time.

\begin{figure*}[!h]
\begin{tcolorbox}[
  colback=palepurple,
  colframe=black!30,
  boxrule=0.5pt,
  arc=2mm,
  left=4pt,
  right=4pt,
  top=2pt,
  bottom=1pt,
  title={\small\textbf{Structured Zero-Shot Prompt}}
]
\begin{lstlisting}[basicstyle=\ttfamily\scriptsize,breaklines=true,frame=none,aboveskip=0pt,belowskip=0pt]
<role>
You are an expert in identifying the direct cause of events from textual evidence.
</role>

<task>
Given an event, context documents, and candidate explanations, analyze systematically to identify the most plausible direct cause(s).
</task>

<input_format>
<context_documents>
<document_1>: [document content]</document_1>
<document_2>: [document content]</document_2>
[additional documents as needed...]
</context_documents>

<target_event>[event description]</target_event>

<options>
<option_a>[option A]</option_a>
<option_b>[option B]</option_b>
<option_c>[option C]</option_c>
<option_d>[option D]</option_d>
</options>
</input_format>

<instructions>

<reasoning_criteria>
- Base your reasoning ONLY on evidence from the provided context documents
- Look for direct causal relationships, not just correlations or temporal sequences
- Test logical sufficiency: Would this factor alone reasonably be enough to cause the event?
- Require both conditions: Direct textual support AND logical sufficiency to cause the event
- Use single-step reasoning: Avoid multi-step causal chains or indirect relationships
- Prioritize explicit causal language: "caused by," "resulted from," "led to," "triggered by," "due to"
</reasoning_criteria>

<selection_rules>
- Multiple options can be correct - choose ALL that apply
- Select multiple options only if each cause has strong evidence and is individually sufficient
- If options contradict each other, select the one with stronger textual evidence
- Always output ALL correct options, including duplicates: if options are duplicate/identical but correct, include both letters
- If none seems perfectly sufficient, select the single best-supported among A-D.
- NEVER create options beyond A, B, C, D
- There is always at least one correct option from A-D
</selection_rules>

<quality_checks>
- Verify each selected option has direct quotes or paraphrases from context
- Ensure you haven't made assumptions beyond what's explicitly stated
- Confirm logical sufficiency: could this realistically cause the event by itself?
- Valid answers in the <answer> section are only A,B,C,D; never output anything else; if uncertain, pick the best-supported among A-D and output it without explanations.
</quality_checks>

</instructions>

<output_format>

Provide your answer in EXACTLY this format (no additional text before or after):

<analysis>
Option A: [Your brief reasoning for option A - 1-2 sentences]
Option B: [Your brief reasoning for option B - 1-2 sentences]
Option C: [Your brief reasoning for option C - 1-2 sentences]
Option D: [Your brief reasoning for option D - 1-2 sentences]
</analysis>

<answer>
[Letter(s) ONLY - e.g., "B" or "B,D" or "C"]
</answer>

CRITICAL FORMATTING RULES:
- Start your response with <analysis> (no text before it)
- End your response with </answer> (no text after it)
- In <answer> tags, write ONLY letters: A, B, C, or D (comma-separated for multiple)
- DO NOT write "Option A" or "Option B" in the <answer> tags - just the letter(s)

</output_format>
\end{lstlisting}
\end{tcolorbox}
\captionof{figure}{Structured zero-shot prompt for direct inference (Section~\ref{sec:inference}). 
}
\label{fig:prompt_direct}
\end{figure*}

\subsection{Direct Prompting}
\label{sec:prompt_direct}

Our primary inference prompt employs a structured analysis-before-answer format that requires the model to evaluate each candidate option independently before committing to a final answer.

\subsection{GEPA Optimization Insights}
\label{sec:gepa_insights}

We employ GEPA~\cite{agrawal2025gepa} via DSPy~\cite{khattab2024dspy} to explore the prompt design space through reflective prompt optimization.
Rather than deploying the resulting prompts directly, we analyze the heuristics GEPA automatically discovered to extract transferable design principles for causal reasoning that informed our final manually-designed prompt (Figure~\ref{fig:prompt_direct}).

\paragraph{Optimization Setup.}
The optimization signature provides the target event, context documents, and candidate options as inputs, with reasoning and answer as outputs.
We intentionally kept the baseline signature minimal(a single docstring with no engineered heuristics) to give GEPA maximum room for discovery.
The metric rewards exact match (1.0) and partial match (0.5), with diagnostic feedback identifying failure types (under-selection, over-selection, wrong answer) to guide reflective evolution.

\paragraph{Practical Observations.}
Our experience with GEPA yields several practical observations for prompt optimization in general.
GEPA is most effective as a \textit{bootstrapping} tool: given no starting prompt, it can rapidly produce a functional one, making it well-suited for agent retraining when adapting to new data distributions or edge cases.
However, it can only mildly optimize an already well-functioning prompt, and optimization quality depends heavily on the pre-optimization starting point.
We also observe that prompt evolution plateaus at the knowledge boundary of the reflection LLM: the optimizer cannot discover reasoning strategies that the reflection model itself does not possess.
Data contamination, a common concern with automatic prompt optimization, can be mitigated by adjusting the reflection prompt to discourage memorization of training examples.
Finally, we note that incorporating an explicit optimization-step directive within the reflection prompt (e.g., instructing the reflector to propose targeted modifications rather than wholesale rewrites) improves stability and convergence across runs.

\paragraph{Discovered Heuristics.}
The best evolved prompt independently surfaced five causal reasoning heuristics without any manual engineering:

\begin{enumerate}[noitemsep,topsep=2pt]
    \item \textbf{Path length prioritization:} Strong causes are 1-2 steps from the target; distal causes are treated as background context unless they meet an exception criterion.
    \item \textbf{Confidence thresholds:} High-confidence causal links are ``almost always'' strong; moderate links require supporting causal language; weak links are generally rejected.
    \item \textbf{Directionality and sibling detection:} Effects of the target are distinguished from causes; shared-root events are classified as side effects rather than causes.
    \item \textbf{Evidence-cause distinction:} Media reports about an event are separated from the event itself as causal triggers.
    \item \textbf{Duplicate handling:} Identical options describing the same event must both be selected.
\end{enumerate}

These heuristics are notable because they emerged purely from optimization pressure on answer correctness, without any explicit instruction about causal theory.

\paragraph{Dataset Bias Discovery.}
Beyond causal reasoning heuristics, GEPA independently surfaced two exploitable artifacts of the dataset construction process.
First, 22.9\% of questions contain \textit{duplicate options} (identical text in different positions) and GEPA learned that selecting one without the other always incurs a scoring penalty, converging on the invariant that identical options must share truth values.
Second, 14.6\% of questions include a ``None of the others are correct causes'' option, and GEPA discovered that co-selecting it with other candidates always produces incorrect answers, effectively learning mutual exclusivity without being told.
These patterns are significant because they emerged purely from optimization pressure: GEPA identified them as deterministic ``free score'' invariants before any manual dataset analysis.
These GEPA-surfaced invariants directly motivated the post-hoc consistency heuristics described in Section~\ref{sec:verification_rules}, specifically R1 (None Exclusivity) and R2 (Duplicate Consistency), which together account for 19 corrections and seed the cross-question propagation chain that yields the full $+$5.6~pp improvement.

\paragraph{Error-Informed Refinement.}
After analyzing the hardest questions (Section~\ref{sec:failure_taxonomy}), we created a second-generation prompt that directly addresses the identified failure patterns:

\begin{itemize}[noitemsep,topsep=2pt]
    \item \textbf{Proximal-cause bias}: Replaced the strict proximity rule with a counterfactual necessity test(``Could the target have occurred without this event?'') allowing enabling conditions at any distance to qualify as valid causes.
    \item \textbf{Single-cause default} (under-selection): Added explicit multi-answer guidance requiring re-evaluation of each remaining option after identifying the strongest cause.
    \item \textbf{Causal chain length disagreement}: Distinguished ``proximate triggers'' from ``enabling conditions,'' both treated as valid cause types.
    \item \textbf{Over-selection penalty}: Added a precision warning noting that incorrect extra selections result in zero score.
\end{itemize}

\paragraph{From GEPA to Final Prompt.}
We deliberately curated which GEPA discoveries to incorporate into the final prompt (Figure~\ref{fig:prompt_direct}).
Structural heuristics that generalize across domains were retained: single-step reasoning preference (reasoning criterion~5), explicit causal language prioritization (criterion~6), duplicate option handling (selection rule~4) and the analysis-before-answer format that GEPA consistently preserved across all evolved prompts.
Conversely, domain-specific patterns that GEPA discovered, such as ``Request $\to$ Approval'' and ``Crisis $\to$ Response'' causal templates, were deliberately excluded to avoid overfitting to the optimization set's topical distribution.
Numeric confidence thresholds were abstracted into the general logical sufficiency test (criterion~3).
The dataset-specific invariants (duplicate consistency, ``None'' exclusivity) were implemented as post-hoc heuristics rather than prompt instructions, since they are deterministic properties better enforced programmatically than via natural-language instruction.


\section{Error Analysis Details}
\label{sec:error_appendix}

\begin{figure*}[t!]
\centering
\includegraphics[width=0.75\textwidth]{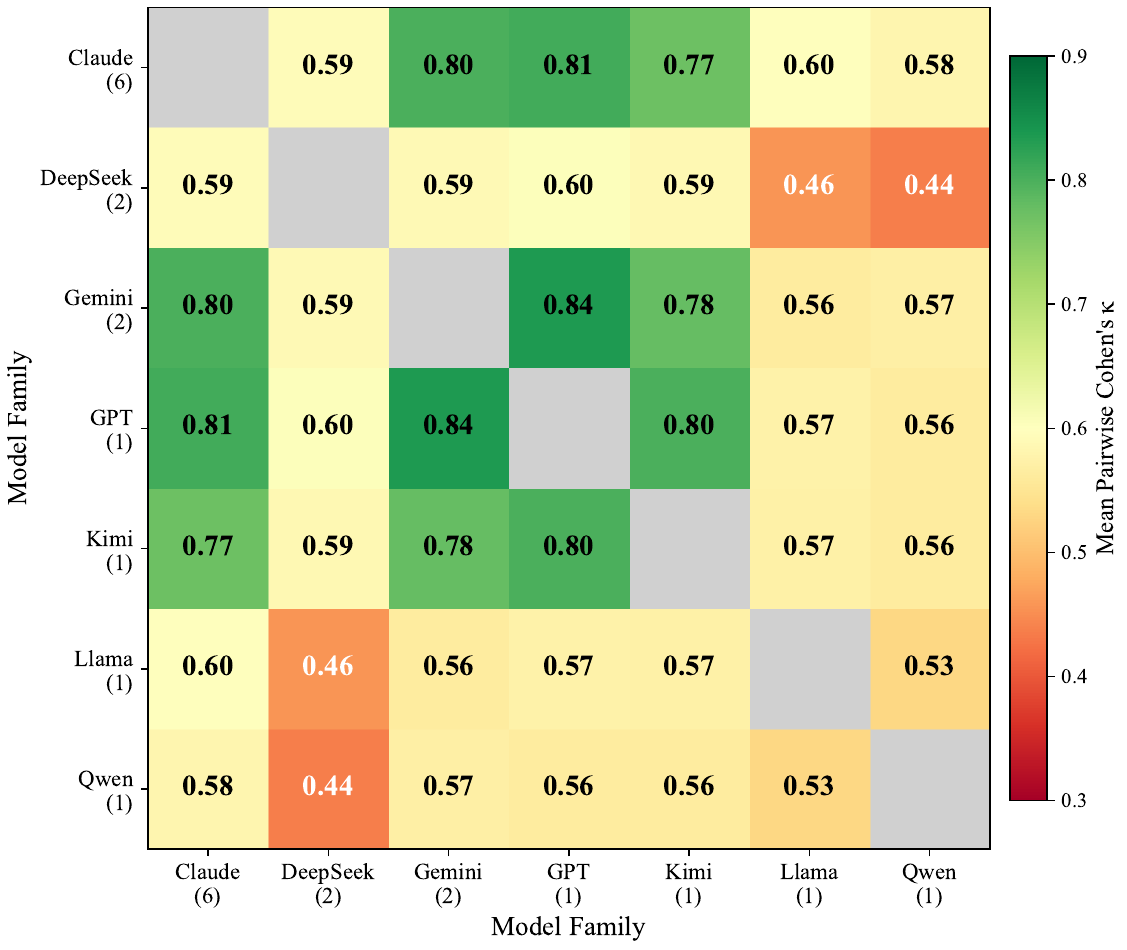}
\caption{Family-level mean pairwise Cohen's~$\kappa$ across 14 models (7~families). Frontier families (Claude, Gemini, GPT, Kimi) show high mutual agreement ($\kappa > 0.77$); DeepSeek is dragged down by R1's low accuracy. Numbers in parentheses = model count per family. Diagonal masked (within-family comparison undefined for single-model families).}
\label{fig:agreement_heatmap}
\end{figure*}

This appendix provides extended analysis supporting the findings in Section~\ref{sec:error_analysis}.
We analyze inter-annotator agreement across 14 models from 7~families, all evaluated on the full dev set (400 questions).
Table~\ref{tab:iaa_metrics} summarizes global agreement metrics and Figure~\ref{fig:agreement_heatmap} shows family-level mean pairwise Cohen's~$\kappa$, revealing that frontier families (Claude, Gemini, GPT) form a high-agreement cluster ($\kappa > 0.78$) while mid-range models (Llama, Qwen) agree less with frontier systems ($\kappa \approx 0.55$).

\begin{table*}[t!]
\centering
\small
\renewcommand{\arraystretch}{1.15}
\begin{tabular}{@{}clllrll@{}}
\toprule
\textbf{QID} & \textbf{Topic} & \textbf{Gold} & \textbf{Consensus} & \textbf{Fail} & \textbf{Multi?} & \textbf{Pattern} \\
\midrule
\multicolumn{7}{@{}l}{\textit{Truly unbreakable: all 14 models score zero (all 7 families fail)}} \\
\midrule
q-2261 & Afghanistan & A,C & D & 14/14 & Yes & Under-sel.\ + proximal \\
q-2356 & DeepSeek & D & C & 14/14 & No & Consensus wrong (7/14 C) \\
q-2411 & DOGE/Musk & A & D & 14/14 & No & Consensus wrong (13/14 D) \\
\midrule
\multicolumn{7}{@{}l}{\textit{Near-unbreakable: $\geq$12 models score zero}} \\
\midrule
q-2341 & Afghanistan & B,C,D & A & 13/14 & Yes & Under-sel.\ + wrong opt. \\
q-2028 & Credit Suisse & A & D & 13/14 & No & Distal-cause preference \\
q-2182 & Afghanistan & C & D & 12/14 & No & Proximal-cause bias \\
q-2225 & Wagner & C & A & 12/14 & No & Proximal-cause bias \\
q-2149 & Amazon fires & C & B & 12/14 & No & Annot.\ ambiguity \\
q-2036 & Myanmar & B & A & 12/14 & No & Distal-cause preference \\
\bottomrule
\end{tabular}
\caption{Hardest questions across 14 models (7~families). Top: 3 questions where all models score zero. Bottom: 6 near-unbreakable questions where $\geq$12 models fail. \emph{Fail} = models scoring zero. \emph{Under-sel.} = multi-answer under-selection; \emph{Consensus wrong} = majority agrees on incorrect option; \emph{Proximal-cause bias} = models prefer temporally nearest cause~\cite{joshi2024llms}.}
\label{tab:unbreakable}
\end{table*}

\subsection{Failure Taxonomy}
\label{sec:failure_taxonomy}

Table~\ref{tab:unbreakable} lists the 3 truly unbreakable questions (all 14 models score zero) and 6 near-unbreakable questions ($\geq$12 models score zero). Four distinct failure patterns emerge:

\paragraph{Proximal-Cause Bias.}
Question q-2182 asks for the cause of ``Taliban fighters spread across Kabul streets.''
The gold answer is C (``US troops began withdrawing from Afghanistan in early July''), but 12/14 models select D (``Afghan President Ashraf Ghani fled the country'').
Models apply a temporal proximity heuristic~\cite{joshi2024llms}, preferring the most recent antecedent, over enabling conditions.

\paragraph{Under-Selection + Wrong Option.}
Question q-2341 (gold: B,C,D) sees 11/14 models selecting only A, the sole \emph{incorrect} option, indicating that models not only under-select but converge on the wrong cause entirely when the gold answer is a large multi-answer set.

\paragraph{Annotation Ambiguity.}
Question q-2149 (Amazon fires, gold: C, 12/14 fail) involves cases where the context supports multiple reasonable interpretations; model disagreement may reflect genuine annotation uncertainty rather than reasoning failure.

\subsection{Topic-Level Difficulty}
\label{sec:topic_difficulty}

\begin{figure*}[t!]
\centering
\includegraphics[width=\textwidth]{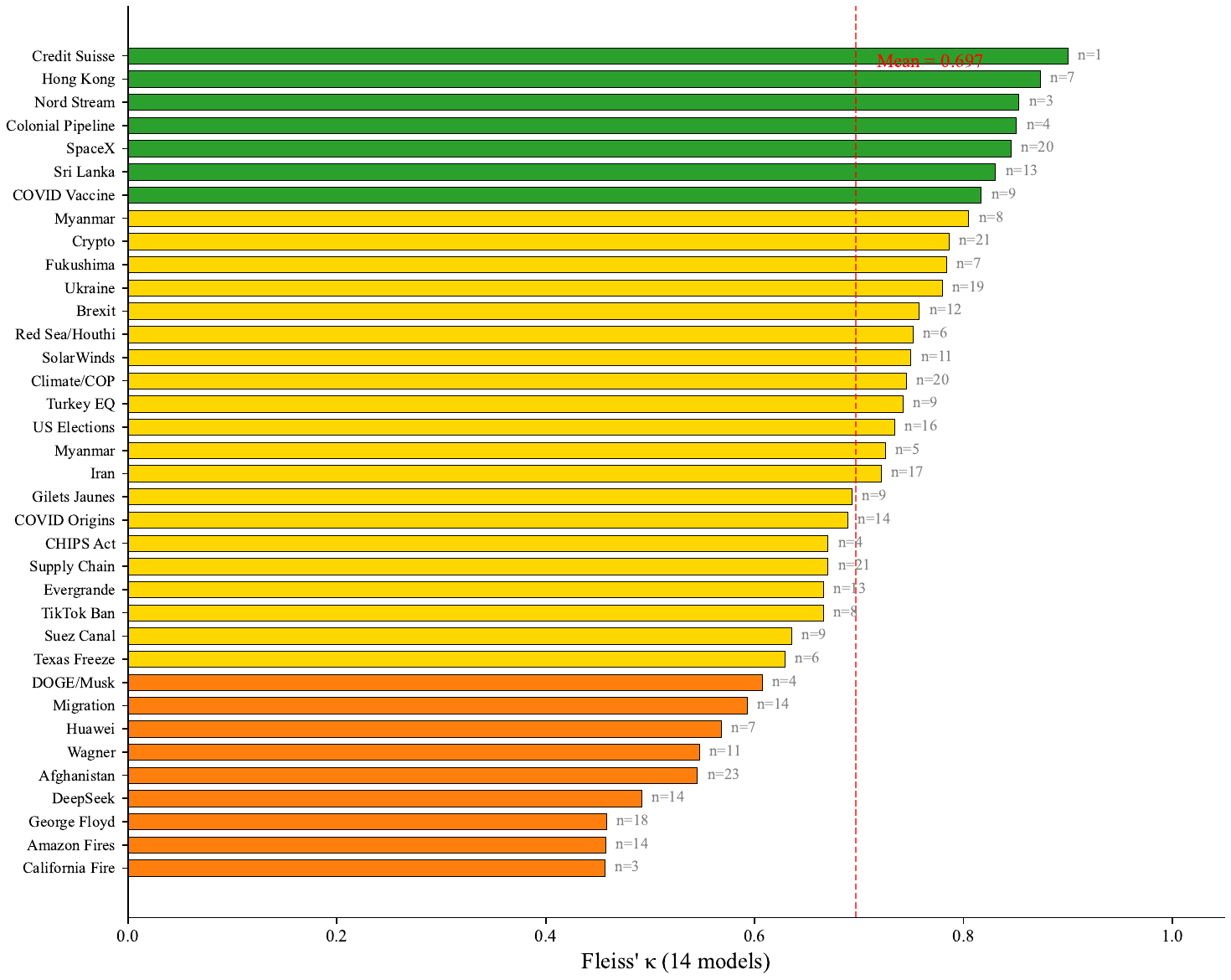}
\caption{Per-topic Fleiss'~$\kappa$ across 14 models (bar labels = question counts per topic); dashed line marks the per-topic mean ($\kappa = 0.697$). Topics are sorted by agreement, coloured by Landis--Koch interpretation: green = almost perfect ($\kappa \geq 0.81$), yellow = substantial ($0.61$--$0.80$), orange = moderate ($0.41$--$0.60$).}
\label{fig:topic_agreement}
\end{figure*}

Table~\ref{tab:topic_difficulty_detailed} reports the hardest topics by exact-match accuracy, alongside structural features that predict difficulty.
Figure~\ref{fig:topic_agreement} shows per-topic Fleiss'~$\kappa$ across all 14 models, revealing substantial cross-topic variance (range: 0.46-0.90).

\begin{table}[t]
\centering
\small
\renewcommand{\arraystretch}{1.15}
\begin{tabular}{@{}clrrr@{}}
\toprule
\textbf{ID} & \textbf{Domain} & \textbf{N} & \textbf{Multi\%} & $\boldsymbol{\kappa_{14}}$ \\
\midrule
26 & California Fire & 3 & --- & 0.456 \\
33 & Amazon fires & 14 & 43 & 0.457 \\
30 & George Floyd & 18 & 72 & 0.458 \\
18 & DeepSeek & 14 & --- & 0.492 \\
32 & Afghanistan & 23 & 57 & 0.545 \\
12 & Wagner & 11 & --- & 0.547 \\
\bottomrule
\end{tabular}
\caption{Hardest topics by Fleiss'~$\kappa$ across 14 models. Topics with complex causal structures and high multi-answer prevalence show the lowest agreement.}
\label{tab:topic_difficulty_detailed}
\end{table}

\begin{table}[t]
\centering
\small
\renewcommand{\arraystretch}{1.15}
\resizebox{\columnwidth}{!}{%
\begin{tabular}{@{}lr@{}}
\toprule
\textbf{Metric} & \textbf{Value} \\
\midrule
Fleiss' $\kappa$~\cite{fleiss1971measuring} & 0.690 \\
Kripp.\ $\alpha$ (Jaccard)~\cite{krippendorff2011computing} & 0.645 \\
Kripp.\ $\alpha$ (nominal) & 0.613 \\
Unanimous agreement & 10.0\% \\
Majority agreement & 100\% \\
\midrule
Within-family $\bar{\kappa}$ & 0.794 \\
Cross-family $\bar{\kappa}$ & 0.661 \\
Unbreakable (all score 0) & 3 \\
No exact match & 42 \\
Oracle (best per Q) & 0.895 \\
\bottomrule
\end{tabular}}
\caption{Inter-annotator agreement metrics across 14 models (7~families: Claude, DeepSeek, Gemini, GPT, Kimi, Llama, Qwen) on 400 dev questions.}
\label{tab:iaa_metrics}
\end{table}

\begin{figure}[t!]
\centering
\includegraphics[width=0.75\columnwidth]{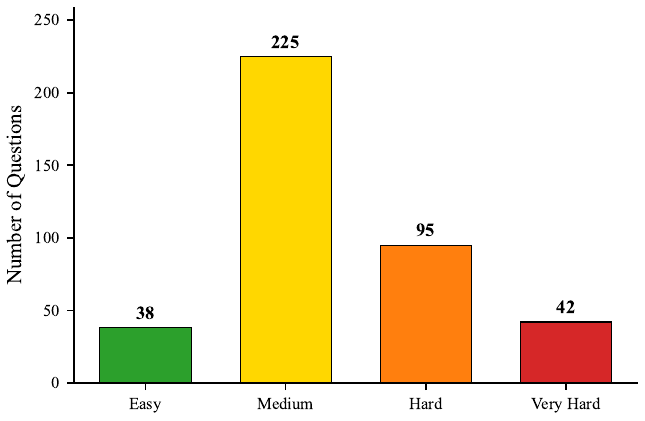}
\caption{Dev-set difficulty distribution across 14 models: Easy 38, Medium 225, Hard 95, Very Hard 42 (based on how many models score zero). The majority of questions are medium-difficulty, with only 42 very hard questions where $\geq$10 models fail.}
\label{fig:difficulty_distribution}
\end{figure}

\begin{table}[t!]
\centering
\small
\renewcommand{\arraystretch}{1.15}
\begin{tabular}{@{}lcrrr@{}}
\toprule
\textbf{Configuration} & $\boldsymbol{k}$ & \textbf{Base} & \textbf{+SC} & $\boldsymbol{\Delta}$ \\
\midrule
\multicolumn{5}{@{}l}{\textit{Same-Model SC ($\tau{=}1.0$)}} \\
\midrule
Claude Sonnet 4.5      & 3 & 0.790 & \textbf{0.806} & $+$1.6 \\
Gemini 3 Flash         & 3 & 0.804 & \textbf{0.810} & $+$0.6 \\
\midrule
\multicolumn{5}{@{}l}{\textit{Multi-Model Ensemble}} \\
\midrule
Son.~Think + GPT + Gemini & 3 & --- & \textbf{0.811} & --- \\
\bottomrule
\end{tabular}
\caption{Self-consistency and multi-model ensemble on the dev set (400 questions). $k$ = number of predictions aggregated via majority voting ($\theta{=}0.5$). Same-model SC provides negligible gains; multi-model combination provides modest gains through model complementarity.}
\label{tab:sc_results}
\end{table}
\begin{table}[t!]
\centering
\small
\renewcommand{\arraystretch}{1.15}
\begin{tabular}{@{}llcc@{}}
\toprule
\textbf{Strategy} & $\boldsymbol{\theta}$ & \textbf{Dev} & \textbf{Test} \\
\midrule
\multicolumn{4}{@{}l}{\textit{Sonnet 4.5 ($k{=}3$, $\tau{=}1.0$)}} \\
\midrule
Baseline ($k{=}1$)       & ---  & 0.790 & 0.904\textsuperscript{\dag} \\
Majority                  & 0.50 & \textbf{0.806} & \textbf{0.902} \\
Union                     & 0.33 & 0.788 & 0.869 \\
Strict Majority           & 0.67 & 0.784 & 0.897 \\
Intersection              & 1.00 & 0.784 & 0.897 \\
\midrule
\multicolumn{4}{@{}l}{\textit{Gemini 3 Flash ($k{=}3$ dev, $k{=}5$ test, $\tau{=}1.0$)}} \\
\midrule
Baseline ($k{=}1$)       & ---  & 0.804 & 0.907 \\
Majority                  & 0.50 & \textbf{0.810} & 0.902 \\
Union                     & 0.33 & 0.788 & 0.888 \\
Strict Majority           & 0.67 & 0.806 & \textbf{0.903} \\
Intersection              & 1.00 & 0.806 & 0.900 \\
\bottomrule
\multicolumn{4}{@{}p{0.9\columnwidth}@{}}{\textsuperscript{\dag}\footnotesize Test uses Sonnet 4.5 Thinking ($k{=}3$).}
\end{tabular}
\caption{Self-consistency aggregation strategies on dev (400~questions) and test (612~questions). $\theta$ = per-option inclusion threshold. \textbf{Bold} = best SC strategy per section.}
\label{tab:aggregation_strategies}
\end{table}

\subsection{Under-Selection Mechanism Analysis}
\label{sec:mechanism_analysis}

\begin{figure*}[t!]
\centering
\includegraphics[width=0.75\textwidth]{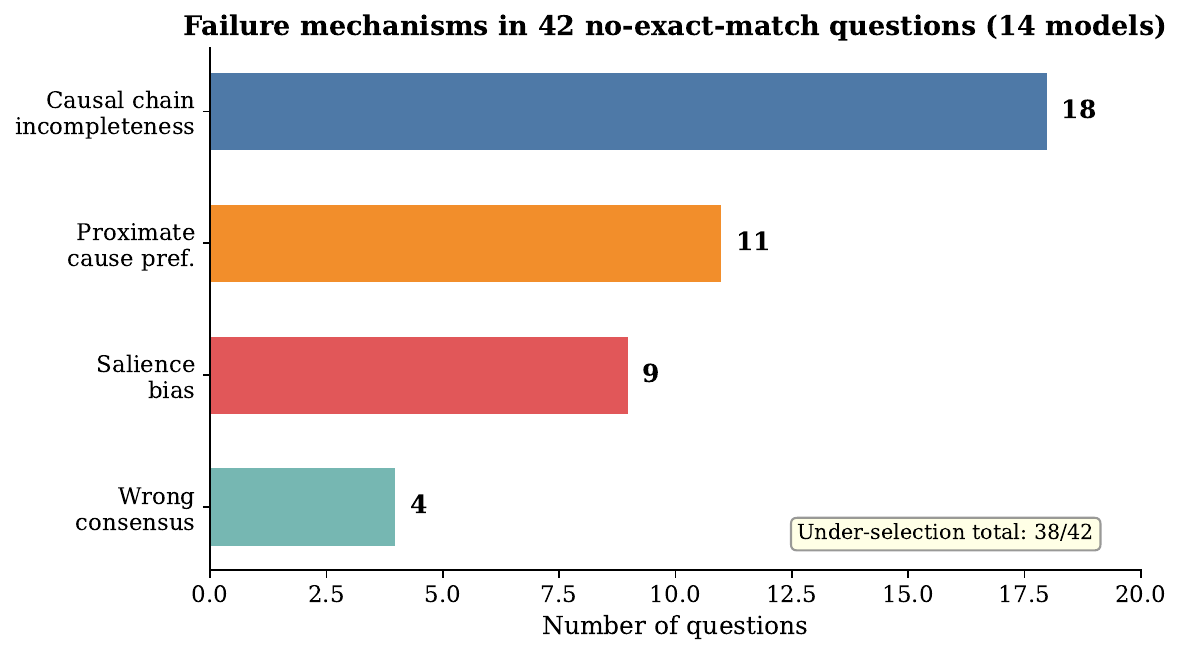}
\caption{Failure mechanism breakdown across 42 no-exact-match questions (14 models). Causal chain incompleteness, proximate cause preference, and salience bias together account for 38/42 failures, all manifestations of the single-cause default bias.}
\label{fig:failure_mechanism}
\end{figure*}

Across the 42 no-exact-match questions (14 models), models select on average 1.2 options when gold requires 2.4, a 51\% cause-count reduction. In 83\% of cases, the consensus prediction is a single option.
Three mechanisms drive under-selection (Figure~\ref{fig:failure_mechanism}):

\paragraph{Causal chain incompleteness (18/42 questions).}
When the gold answer spans multiple links in a causal chain (root cause $\to$ intermediate events $\to$ target), models select only one link and omit the rest~\cite{chi2024unveiling, yu2025causaleval}.
For example, q-2135 (Korean martial law) has gold=\{A,C,D\} covering the full chain (declaration $\to$ soldiers stormed $\to$ annulment $\to$ impeachment), yet all 14 models select only~A.
The pattern recurs across the Capitol riot questions (q-2044, q-2094, q-2252, q-2299), where models consistently miss intermediate steps in the protest $\to$ breach $\to$ lockdown $\to$ halt sequence.

\paragraph{Proximate cause preference (11/42 questions).}
Models favour the most causally proximate antecedent over distal enabling conditions~\cite{joshi2024llms}, whether the distance is temporal (months between events) or structural (enabling process vs.\ its outcome).
This is most severe in the wrong-answer failures: in q-2261 (Afghanistan), all 14 models choose ``Ghani fled'' (proximal trigger) over ``US withdrawal'' (enabling condition), scoring zero.
The same pattern recurs across five Brexit questions (q-2090, q-2210, q-2233, q-2286, q-2322), where models pick the vote outcome (``voted to leave'') but miss the enabling institutional act (``held a referendum''), events on the same day, distinguished by causal role rather than temporal order.

\paragraph{Salience bias (9/42 questions).}
Models prefer dramatic, newsworthy causes over subtler contributing factors~\cite{koo2024benchmarking}.
In q-2226 (Beirut explosion), all 14 models select ``explosion killing 200 people'' but miss ``destroyed medical infrastructure'': both are valid causes of the PM's resignation, but models anchor on the more dramatic event.

Multi-answer prevalence correlates positively with topic difficulty (Spearman $\rho > 0$) across all 36 topics.
Topic~33 (Amazon fires, $\kappa = 0.457$) reflects cyclical causation where deforestation causes fires, fires trigger policy responses, and policy failures enable further deforestation.
Topic~30 (George Floyd/BLM, $\kappa = 0.458$, 72\% multi-answer) shows how social movement causation creates branching chains where many options are partially valid.

\begin{figure*}[t!]
\centering
\includegraphics[width=0.85\textwidth]{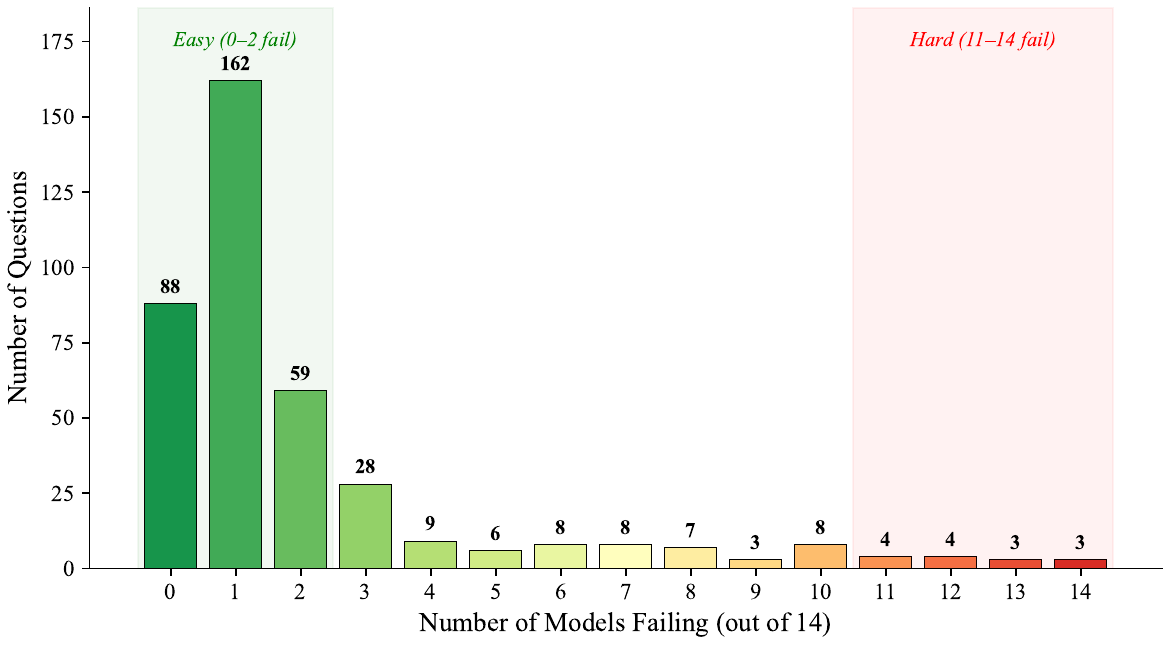}
\caption{Failure gradient across 14 models: distribution of how many models fail each question. 88 questions (22\%) are solved by all 14 models; only 3 questions (0.75\%) defeat all 14. The long right tail (10-14 failures) contains 22 questions requiring systematic improvements.}
\label{fig:failure_gradient}
\end{figure*}

\begin{figure*}[t!]
\centering
\includegraphics[width=0.85\textwidth]{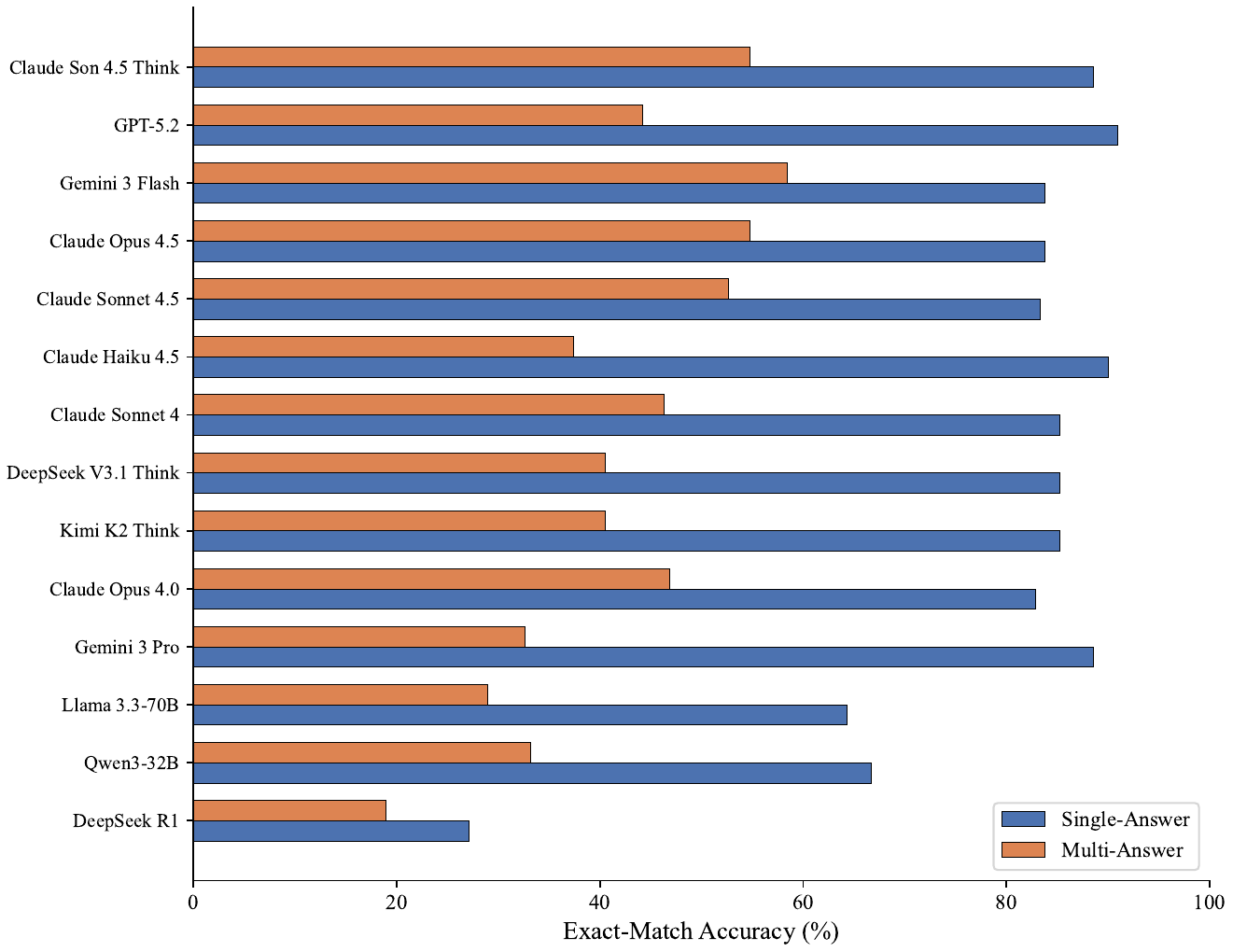}
\caption{Per-model exact-match accuracy by answer cardinality across 14 models. Every model shows a single-to-multi gap (25-56~pp); under-selections (1,389 total) vastly exceed over-selections (52), confirming conservative cause selection as the dominant shared bias.}
\label{fig:model_accuracy}
\end{figure*}

\subsection{Model Agreement and Ensemble Analysis}
\label{sec:ensemble_appendix}

\paragraph{Inter-Annotator Agreement.}
Table~\ref{tab:iaa_metrics} reports agreement metrics across 14 models.
Fleiss'~$\kappa = 0.690$ (\emph{substantial}) reflects the inclusion of weaker models (DeepSeek~R1, Llama, Qwen) that increase prediction variance.
Within-family agreement ($\bar{\kappa} = 0.794$) exceeds cross-family agreement ($\bar{\kappa} = 0.661$), confirming that architecturally similar models share prediction patterns.
Agreement is higher on single-answer questions ($\kappa_{14} = 0.699$) than multi-answer questions ($\kappa_{14} = 0.668$), consistent with the inherently greater difficulty of multi-answer evaluation.

\paragraph{Ensemble Analysis.}
The oracle upper bound of 0.895 (selecting the best model per question) indicates 6.7~pp headroom from model complementarity, with unique correct predictions distributed across families: Gemini~Flash~7, Claude~(family total)~9, Llama~3, GPT-5.2~2.
Bootstrapped significance tests yield $p > 0.05$ for all pairwise model comparisons, indicating that observed differences are not statistically significant at $\alpha = 0.05$ given the 400-question evaluation set, a limitation inherent to shared-task evaluations.

\subsection{Self-Consistency and Ensemble Analysis}
\label{sec:aggregation_strategies}

Table~\ref{tab:sc_results} reports the impact of self-consistency sampling and multi-model ensembling on the dev set.
Same-model self-consistency ($k{=}3$, $\tau{=}1.0$) provides modest accuracy gains: Sonnet~4.5 improves by 1.6~pp and Gemini~3 Flash improves by 0.6~pp, suggesting that sampling variation at temperature~1.0 benefits causal reasoning moderately.
In contrast, majority voting across three architecturally diverse systems (Claude Sonnet~4.5 Thinking, GPT-5.2, Gemini~3 Flash) yields 0.811, exceeding any same-model SC configuration.

Table~\ref{tab:aggregation_strategies} compares aggregation strategies across both models and both splits.
Four per-option inclusion thresholds are evaluated: union ($\theta{=}0.33$, include if any sample selects), majority ($\theta{=}0.50$), strict majority ($\theta{=}0.67$), and intersection ($\theta{=}1.0$, all samples must agree).
With $k{=}3$, strict majority and intersection yield identical results because no option can exceed $\frac{2}{3}$ without unanimous agreement; the difference emerges with $k{=}5$ (Gemini test).

Majority voting ($\theta{=}0.5$) is the most robust strategy, matching or marginally improving baseline across all configurations.
Union ($\theta{=}0.33$) consistently reduces accuracy by 1-4~pp across all configurations due to over-selection, particularly on the test set where single-answer prevalence is higher.
Strict thresholds ($\theta \geq 0.67$) under-select on the dev set, where 47.5\% of questions require multi-label answers; with $k{=}3$ these thresholds reduce to unanimous agreement and yield identical results.


\end{document}